\documentclass[sigconf]{acmart}
\pdfoutput=1
\usepackage{graphicx}
\usepackage{amsmath}
\usepackage{booktabs}
\usepackage{float}
\usepackage{mathtools}
\usepackage{algorithm}
\usepackage{algorithmic}
\usepackage{enumitem}
\usepackage{lipsum}
\usepackage{hyperref}

\AtBeginDocument{%
  }

\copyrightyear{2023} 
\acmYear{2023} 
\setcopyright{acmlicensed}
\acmConference[MM '23]{Proceedings of the 31st ACM International Conference on Multimedia}{October 29-November 3, 2023}{Ottawa, ON, Canada}
\acmBooktitle{Proceedings of the 31st ACM International Conference on Multimedia (MM '23), October 29-November 3, 2023, Ottawa, ON, Canada}
\acmPrice{15.00}
\acmDOI{10.1145/3581783.3612588}
\acmISBN{979-8-4007-0108-5/23/10}

\begin{document}

\title{Null-text Guidance in Diffusion Models is \\Secretly a Cartoon-style Creator}

\author{Jing Zhao}
\authornote{Work done during an internship at JD Explore Academy.}
\orcid{0000-0002-0049-1802}

\affiliation{%
  \institution{College of Computer Science and Technology, National University of Defense Technology}
  \city{Changsha}
  \country{China}
  \postcode{410073}
}
\email{zhaojing@nudt.edu.cn}

\author{Heliang Zheng}
\affiliation{%
  \institution{JD Explore Academy}
  \streetaddress{1 Th{\o}rv{\"a}ld Circle}
  \city{Beijing}
  \country{China}}
\email{zhenghllj@gmail.com}

\author{Chaoyue Wang}
\affiliation{%
  \institution{JD Explore Academy}
  \city{Beijing}
  \country{China}
}
\email{chaoyue.wang@outlook.com}

\author{Long Lan}
\affiliation{%
 \institution{College of Computer Science and Technology, National University of Defense Technology}
   \city{Changsha}
  \country{China}
  \postcode{410073}
 }
 \email{long.lan@nudt.edu.cn}

\author{Wanrong Huang}
\affiliation{%
 \institution{College of Computer Science and Technology, National University of Defense Technology}
   \city{Changsha}
  \country{China}
  \postcode{410073}
 }
 \email{huangwanrong12@nudt.edu.cn}

\author{Wenjing Yang}
\authornote{Corresponding author.}
\affiliation{%
 \institution{College of Computer Science and Technology, National University of Defense Technology}
   \city{Changsha}
  \country{China}
  \postcode{410073}
 }
 \email{wenjing.yang@nudt.edu.cn}

\renewcommand{\shortauthors}{Jing Zhao et al.}

\begin{abstract}
Classifier-free guidance is an effective sampling technique in diffusion models that has been widely adopted. The main idea is to extrapolate the model in the direction of text guidance and away from null-text guidance. In this paper, we demonstrate that null-text guidance in diffusion models is secretly a cartoon-style creator, i.e., the generated images can be efficiently transformed into cartoons by simply perturbing the null-text guidance.
Specifically, we proposed two disturbance methods, i.e., Rollback disturbance (Back-D) and Image disturbance (Image-D), to construct misalignment between the noisy images used for predicting null-text guidance and text guidance (subsequently referred to as \textbf{null-text noisy image} and \textbf{text noisy image} respectively) in the sampling process. 
Back-D achieves cartoonization by altering the noise level of the null-text noisy image via replacing $x_t$ with $x_{t+\Delta t}$.
Image-D, alternatively, produces high-fidelity, diverse cartoons by defining $x_t$ as a clean input image, which further improves the incorporation of finer image details.
Through comprehensive experiments, we delved into the principle of noise disturbing for null-text and uncovered that the efficacy of disturbance depends on the correlation between the null-text noisy image and the source image. Moreover, the proposed methods, which can generate cartoon images and cartoonize specific ones, are training-free and easily integrated as a plug-and-play component in any classifier-free guided diffusion model.
The project page is available at \url{https://nulltextforcartoon.github.io/}.

\end{abstract}


\begin{teaserfigure}
\vspace{-1em}	
  \includegraphics[width=\textwidth]{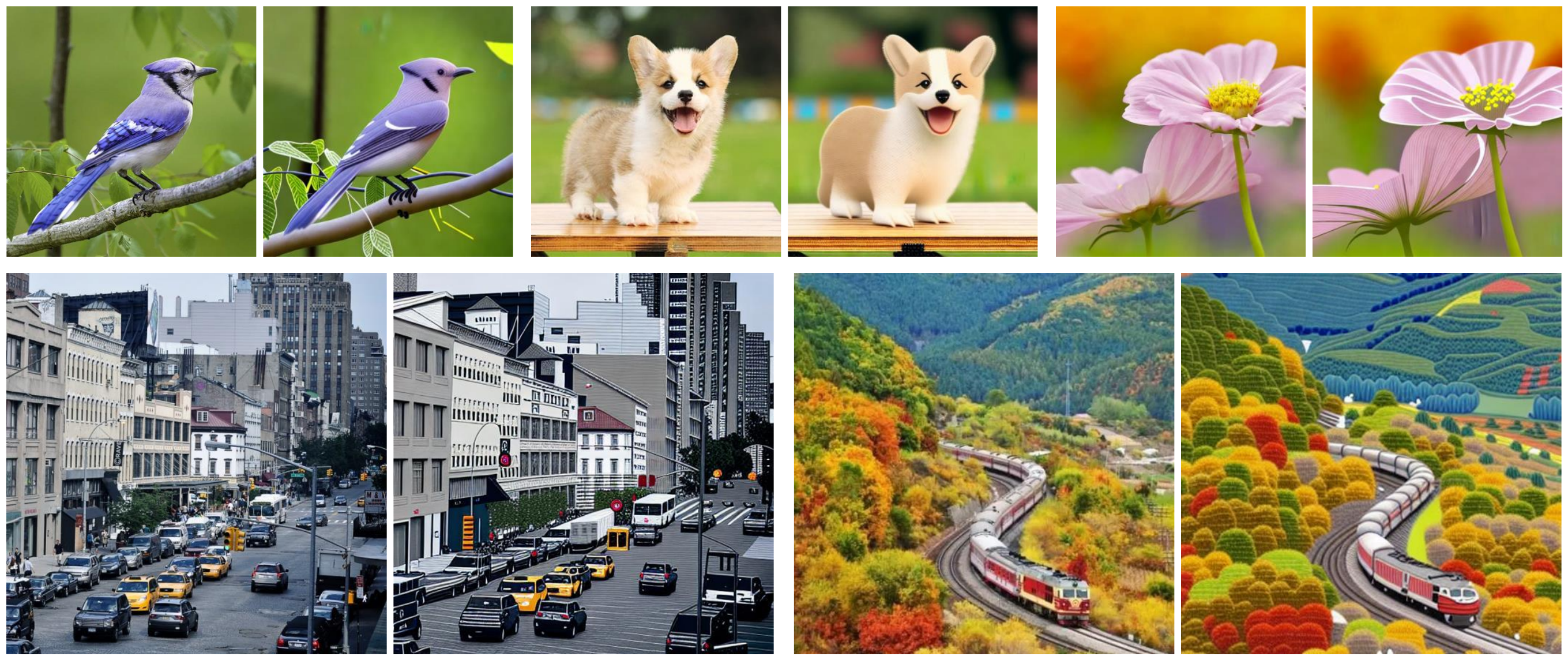}
   \vspace{-2.5em}	
  \caption{Image cartoonization with the proposed Image Disturbance (Image-D) strategy. In each pair, left$\stackrel{cartoonize}{\longrightarrow}$right.
  }
  \vspace{1em}	
  \label{first_page}
\end{teaserfigure}


\maketitle

\section{Introduction}
\label{introduction}

Diffusion models~\cite{HoJA20,DhariwalN21,NicholD21,SongME21,WatsonCH022,abs-2206-00386,RombachBLEO22,ChenCW21} have recently emerged as a compelling topic in computer vision and have shown remarkable results in the field of generative modeling~\cite{SongE19,DicksteinW15,HoJA20}, which generate samples by gradually removing noise from a signal with the training objective expressed as a reweighted variational lower-bound~\cite{HoJA20}. 
Based on diffusion models~\cite{HoJA20,0011SKKEP21,NicholD21}, Dhariwal et al. proposed classifier guidance~\cite{DhariwalN21}, which boosts sample quality but requires an extra trained classifier. 
Classifier-free guidance~\cite{classifier_free} is an alternative technique that modifies predict noise without a classifier by extrapolating the model in the direction of text guidance and away from null-text guidance. 
This technique has significantly improved the image generation capability in diffusion models such that it has gained wide adoption in recent works~\cite{zhao2023magicfusion, Chitwan22,RombachBLEO22}. 
 
To investigate the influence of null-text guidance in classifier-free guidance, we introduced a misalignment between the noisy images used for predicting null-text guidance and text guidance (subsequently referred to as \textbf{null-text noisy image} and \textbf{text noisy image} respectively) by employing noise disturbance anchored by text guidance. Our findings indicate that null-text guidance in diffusion models is secretly a cartoon-style creator.
Specifically, we propose two noise disturbance methods, i.e., Rollback disturbance (Back-D) and Image disturbance (Image-D), to achieve the misalignment of noisy images. Back-D involves modifying the noise level of the null-text noisy image by replacing $x_t$ with $x_{t+\Delta t}$,  resulting in a final output resembling a cartoon. On the other hand, Image-D uses a given clean image as the null-text noisy image, enabling the capture of finer image details and producing high-fidelity cartoon images with increased diversity.

We systematically investigated and analyzed the impact of various hyper-parameters on the proposed methods, elucidating the appropriate conditions for effective cartoonization. 
Our exploration discovered the following insights: 
1) The null-text noisy image that is more closely related to the input image can lead to improved cartoonization outcomes.
2) Text guidance can enhance the diversity and creativity of generated cartoons due to the image-generating capacity inherent in the diffusion model.
3) For effective cartoonization, the noise disturbance needs to form a stable direction for image generation.

The experimental findings demonstrate that the proposed methods facilitate the generation of cartoon depictions encompassing portraits, animals, landscapes, and architectures, among other entities, whilst also enabling the creative cartoonization of specific images. Remarkably, our approach obviates the necessity for training, thereby facilitating simplified implementation in a classifier-free guided diffusion model. 
In summary, this work makes pioneering contributions in the following ways:
\begin{itemize}
\item We conducted an in-depth exploratory analysis of null-text guidance and discovered that null-text guidance is secretly a cartoon-style creator.
\item we put forth plug-and-play cartoonization components, including Rollback disturbance (Back-D) and Image disturbance (Image-D), that enable the free generation of cartoons as well as the cartoonization of specific input images.
\item Our methodology reveals the underlying principles of null-text guidance, thereby contributing further understanding of the principles and potential applications of classifier-free guidance.
\end{itemize}

\begin{figure*}[h]
  \centering
  \includegraphics[width=0.95\linewidth]{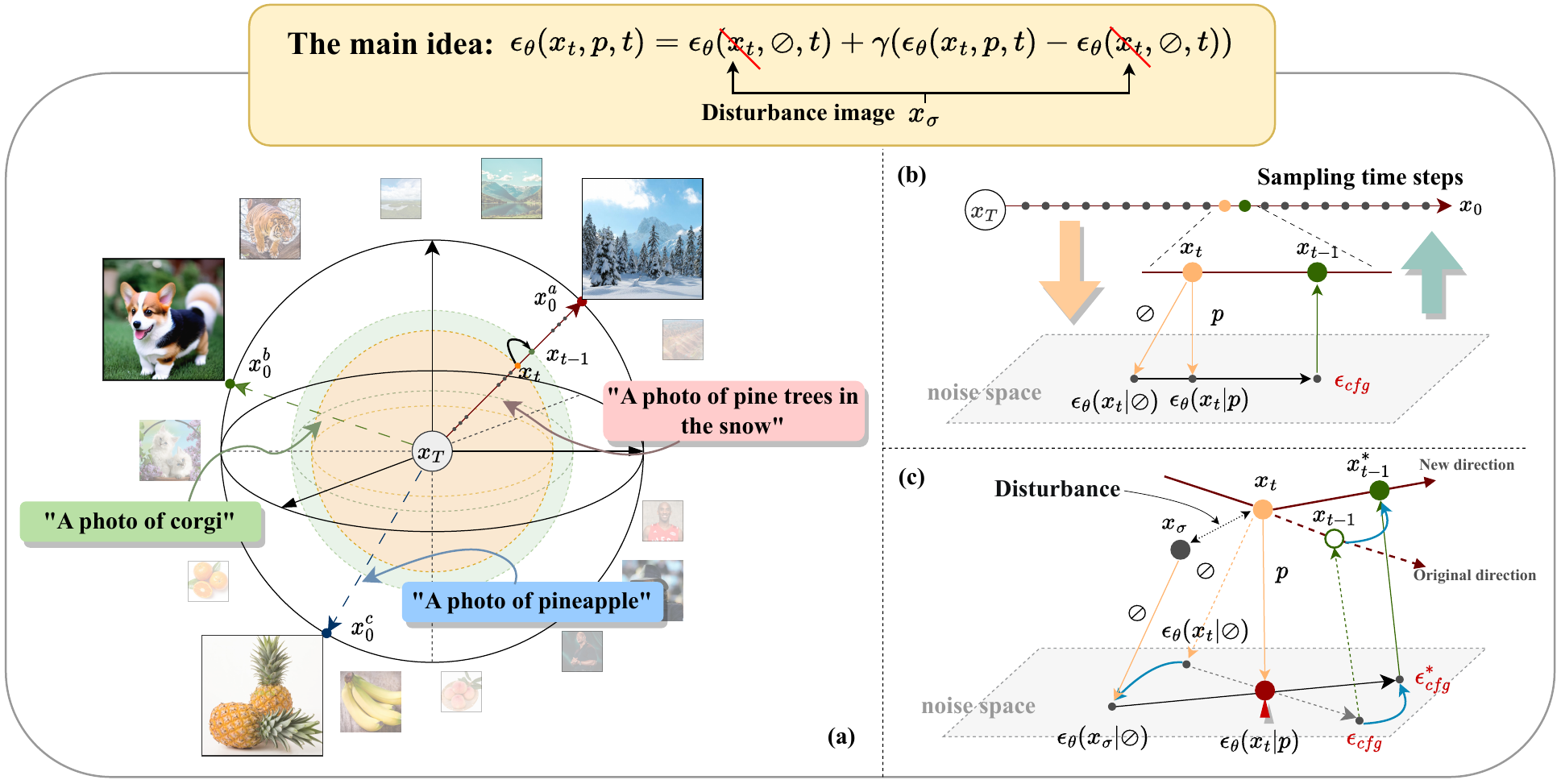}
  \vspace{-1em}
  \caption{Conceptual diagram of the proposed methods. (a)
  $x_T$ is random noise in the center of the image space, which is gradually pushed toward the specific image out by a sampling process guided by a given prompt.
  (b) 
  Predicted noise $\epsilon_{cfg}$ in classifier-free guidance. 
  (c) 
  The proposed Noise Disturbance replaced the null-text noisy image with $x_{\sigma}$, resulting in the change in the direction of generation.
  $\epsilon_{cfg}^{*}$ represents the predicted noise after disturbance. 
  }
  
  \Description{}
  \vspace{-2mm}
\label{fig_method}
\end{figure*}

\section{Related works}
\label{Relatedworks}

\subsection{Classifier-free guidance}
Classifier-free guidance~\cite{classifier_free} is a powerful sampling technique as it directs the model towards text guidance and away from null-text guidance by introducing a null-text guidance term. 
Compared to the previous study, classifier guidance~\cite{DhariwalN21}, which utilizes a separate classifier to trade off Inception Score (IS) and Fréchet Inception Distance (FID) via truncation or low-temperature sampling, classifier-free guidance can be easily implemented and applied. 
Specifically, classifier-free guidance trains an unconditional denoising diffusion model together with the conditional model and updates the prediction of noise by increasing the distance between target noise and null-text noise. 

The utilization of classifier-free guidance has greatly enhanced the caliber of generated images and has become ubiquitous in subsequent studies~\cite{dreambooth22, Chitwan22, NicholDRSMMSC22, LugmayrDRYTG22, Jonathan22, GafniPASPT22, Rinon22, Bahjat22, Amir22, Guy22, zhao2023magicfusion, li20223ddesigner}.
Based on classifier-free guidance, stable diffusion~\cite{RombachBLEO22} has facilitated the training of diffusion models on restricted computational resources, whilst preserving their quality and adaptability by deploying them within the potent pretrained autoencoder's latent space. This has resulted in marking new milestones in the realm of image inpainting and class-conditional image synthesis whilst exhibiting exceptionally competitive performance across multiple tasks. Considering this, our study endeavors to further explore the potency of classifier-free guidance in conjunction with stable diffusion acting as the cornerstone of our investigation.

\subsection{Image cartoonization}
The art form of the cartoon has gained immense popularity and has been widely utilized across diverse domains. In the field of image synthesis, Generative Adversarial Network (GAN)~\cite{GoodfellowPMXWOCB14, 0001ZXFT16, ZhangCYD21a} is a potent technique that enables the generation of data with the same distribution as that of input data by solving a min-max problem between a generator network and a discriminator network. This approach holds considerable potential in generating images that are seamlessly indistinguishable from real images~\cite{JohnsonAF16, CycleGAN2017, PathakKDDE16, SanakoyeuKLO18, RenLG20, LuppinoKBMSJA22, abs-2207-13184, cartoongan18}. 
White-box Cartoon~\cite{WangY20} also employs a GAN architecture to generate cartoonized images. However, it differs from CartoonGAN by allowing each extracted representation to have its own learning objectives. This makes the framework adjustable and controllable. AnimeGAN~\cite{ChenLC19} proposes a unique method for converting real-world scene photographs into anime-style imagery. The approach fuses neural style transfer and GAN to achieve this transformation rapidly while upholding high-quality standards.

Unlike these GAN-based methods~\cite{cartoongan18, WangY20, ChenLC19, ZhuangY21, cartoonstylegan221}, 
This work presents the incorporation of noise interference occurring in the sampling phase of diffusion models to create cartoon images, utilizing not only the input image but also supplementary textual cues.
\textbf{To the best of our knowledge, this is the first instance of utilizing a diffusion model for implementing image cartoonization without model training.}

\section{Method}
\label{method}
\subsection{Preliminaries}
The success of classifier-free guidance lies in introducing null-text guidance to re-predict the noise output by extrapolating the model in the direction of text guidance and away from null-text guidance. 
Specifically, the noise output in the sampling process of classifier-free guidance is computed as follows:
\begin{equation}\label{cfg}
    \epsilon_{\theta}(x_t| p) = \epsilon_\theta (x_t | \oslash ) + \gamma (\epsilon_{\theta}(x_t| p) - \epsilon_\theta (x_t | \oslash ) ),
\end{equation}
where $\epsilon_{\theta}(x_t|\cdot)$ is a simplified notation for $\epsilon_{\theta}(x_t,\cdot,t)$, $\epsilon_\theta(\cdot)$ represents the prediction noise of the U-Net model~\cite{Unet15} parameterized by $\theta$. $t$ indicates the time step and $p$ is the prompt used to guide the generated content. $\oslash$ represents the null-text and $\gamma$ is the guidance scale used to control the strength of the classifier-free guidance. 
Taking the DDIM sampling for example, a denoising step with $\epsilon_t = \epsilon_{\theta}(x_t| p) $ can be denoted as:
\begin{equation}\label{x_t}
    x_{t-1}=\sqrt{\bar{\alpha}_{t-1}}(\frac{x_t-\sqrt{1-\bar{\alpha} _t}\epsilon_t }{\sqrt{\bar{\alpha}_t} } )+\sqrt{1-\bar{\alpha }_{t-1} }\epsilon_t, 
\end{equation}
where $x_t$ is the noisy image in step $t$. $\bar{\alpha}$ is related to a pre-defined variance schedule. 

Classifier-free guidance greatly improves the generation effect of text-to-image tasks based on the diffusion model. We believe that there are more potential functions in null-text guidance worth exploring.

\begin{figure}[t]
  \centering
  \includegraphics[width=\linewidth]{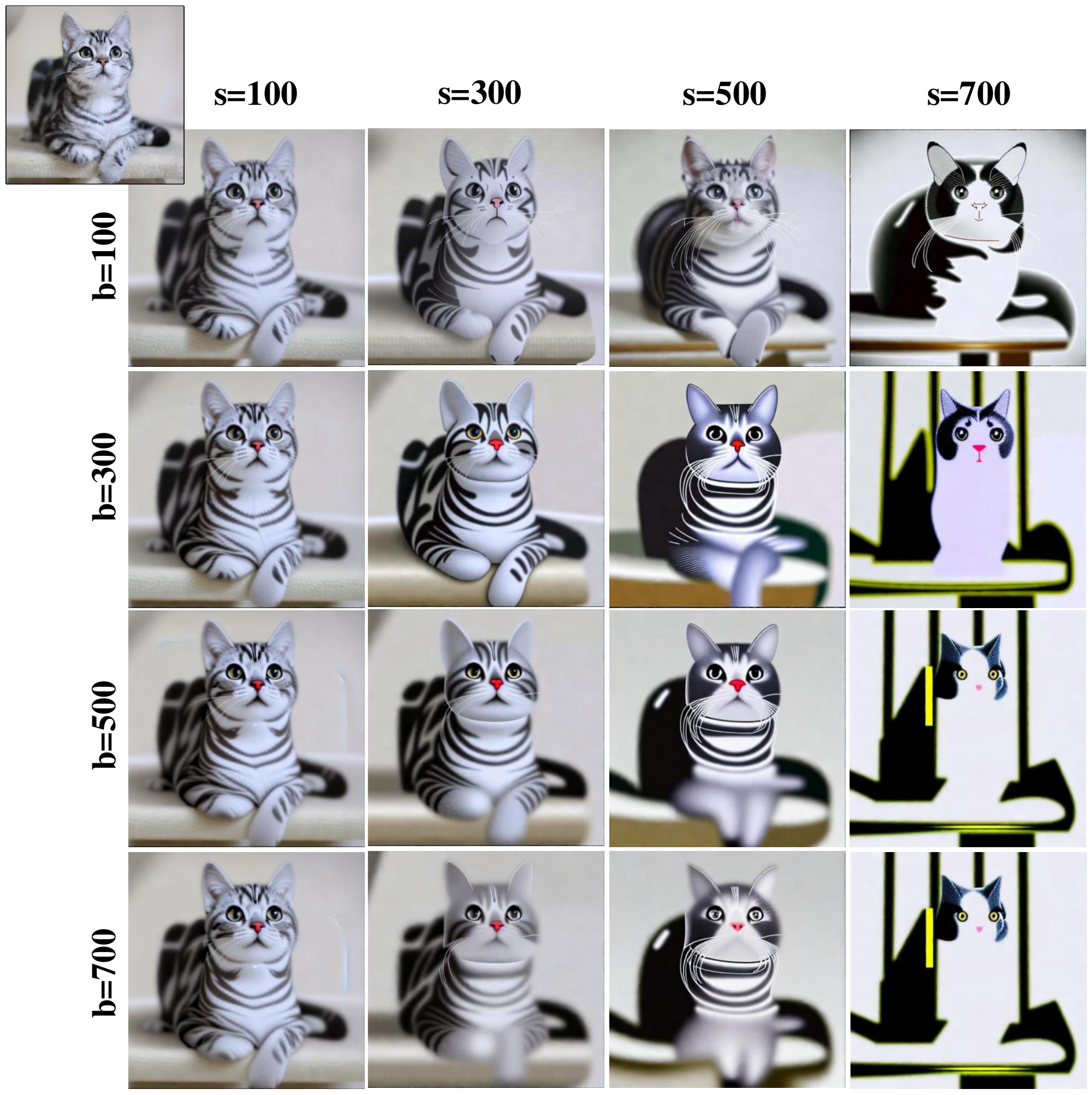}
  \vspace{-2.3em}
  \caption{Exploration of Back-D with rollback step $b$ and disturbance time $s$. The original image $x_0$ located at the top-left corner.
  }
  \label{study_sb}
\end{figure}

\subsection{Noise disturbances
}
The text-guided diffusion model is used to denoise a random noise $x_T$ to a clean image from step $T$ to $0$ with the guidance of the given text prompt as shown in Figure~\ref{fig_method}(a). 
During this procedure, the starting noise $x_T$ is viewed as the central point of the image space, traveling to different endpoints $x_0$ with the assistance of varied prompts following $T$ time samplings.
For the sampling process of classifier-free guidance, the model first predicts the noise guided by text $p$ and null-text $\oslash$ respectively, getting the noises $\epsilon_\theta(x_t|p)$ and  $\epsilon_\theta(x_t|\oslash)$, and then re-compute the noise output as shown in Eq.\eqref{cfg} in the noise space, which is denoted as $\epsilon_{cfg}$ in Figure~\ref{fig_method}(b).
Finally, the noisy image $x_{t-1}$ is obtained by mapping from noise space to the image space via Eq.\eqref{x_t}.

In this work, we seek to investigate further potential functions of null-text guidance by introducing a misalignment between null-text noisy image and text noisy image. 
As illustrated in Figure~\ref{fig_method}(c), a disturbance image $x_{\sigma}$ is chosen from the image space as the null-text noisy image while keeping the text noisy image $x_t$ unchanged, thus forming the noise disturbance anchored by text guidance.
The summit of Figure~\ref{fig_method} manifests this main idea.
As a result, the sampling direction has been altered from $x_t \rightarrow x_{t-1}$ to $x_t \rightarrow x_{t-1}^*$ through the addition of noise disturbance. 



\subsection{Rollback Disturbance (Back-D)}

To conduct the misalignment between null-text noisy image and text noisy image, resulting in the noise disturbance and leading to the change of sampling direction, we propose the Rollback disturbance (Back-D) strategy, which set $x_{\sigma} = x_{t+b}, b \in (0, T),$
where $b$, i.e., $\Delta t$, is a hyper-parameter to control the rollback degree. Compared to $x_t$, $x_{t+b}$ contains more noise and is closer to the initial noisy image $x_T$. 

To preserve the initial structure of the original image, the imposition of noise perturbations is limited to the concluding stage of the sampling procedure. Specifically, such disturbances are initiated solely when $t < s$, where $s$ is a hyper-parameter.
As demonstrated in Figure~\ref{study_sb},
we observed the presence of an ideal cartoon-style version near ($b$=300, $s$=300) and the degree of cartoonization in the generated images is insufficient when $b$ and $s$ are too small (e.g., 100). Conversely, when $b$ and $s$ are excessively large (e.g., 700), the directional deviation caused by noise disturbance is excessive, leading to blurry or even chaotic content in the generated images. In a word, Back-D can enable the free generation of cartoons expeditiously by utilizing a prompt as input. 
A concise summary of the algorithm is presented in Algorithm \ref{algo1}.

\subsection{Image Disturbance (Image-D)}
Inspired by the free generation of cartoons with Back-D, we endeavor to achieve image cartoonization through Back-D. 
Unlike the free generation task that only requires a prompt for guidance, image cartoonization necessitates the use of the base structure of the designated input image, denoted as $x_{ref}$, as the initial noisy image.
Precisely, the process involves obtaining the initial noisy image $x_s$ by adding noise to input image $x_{ref}$ in $s$ steps, wherein the sampling process from $t=s$ to $t=0$ is consistent with that used in free generation.

\begin{figure}[t]
  \centering
  \includegraphics[width=0.9\linewidth]{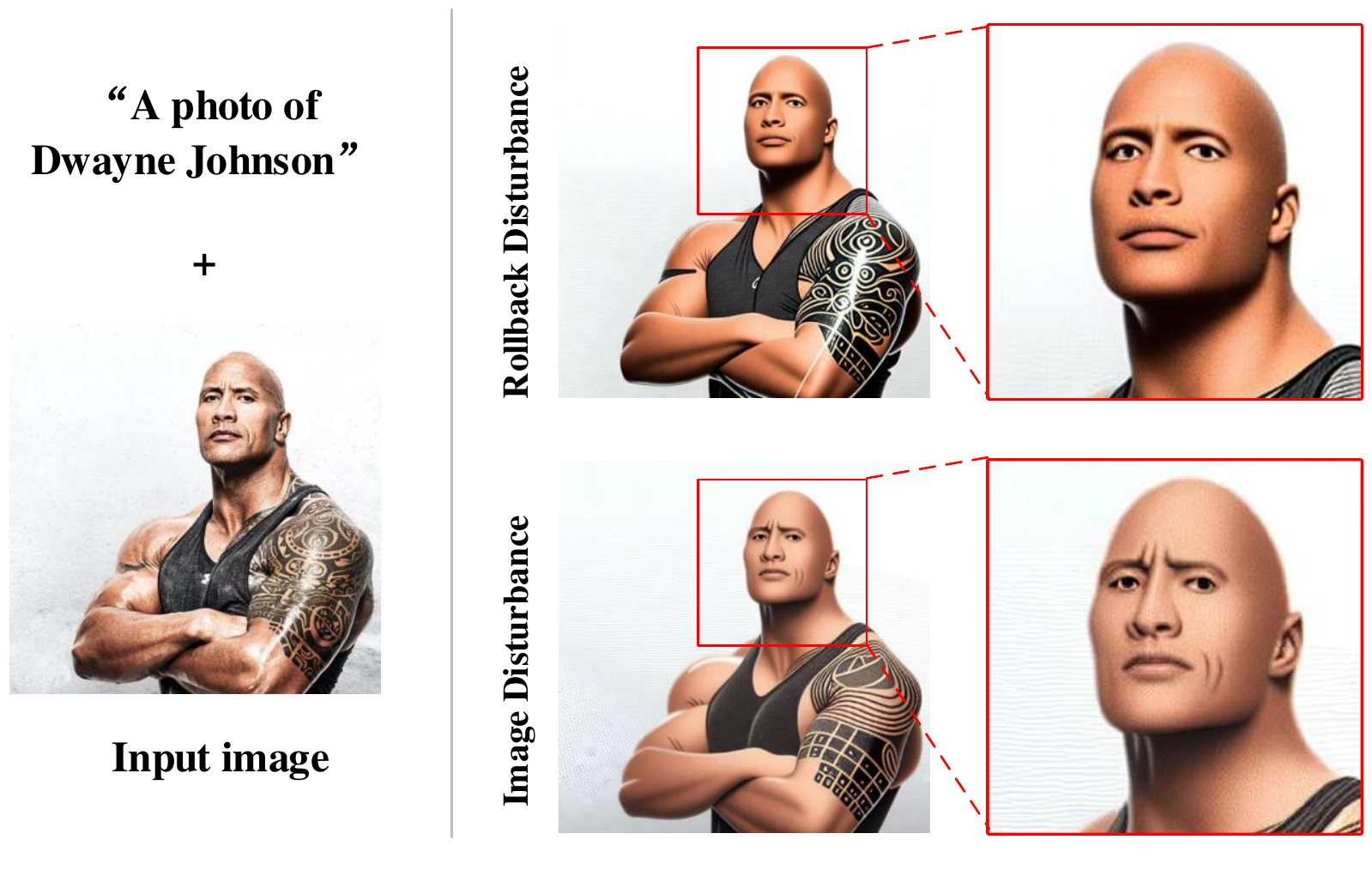}
  \vspace{-1.5em}
  \caption{Comparison between Back-D and Image-D of portrait cartoonization. 
  Image-D can generate better fidelity.
  }
  \label{compare_ri}
\end{figure}

The first row on the right of Figure~\ref{compare_ri} shows the results of Image cartoonization with Back-D, in which the generation is guided by the input image $x_{ref}$ and the prompt "a photo of Dwayne Johnson". 
Despite its ability to preserve the image structure and enable cartoonization for images, Back-D fails to achieve sufficient fidelity.

To enhance the fidelity of image cartoonization, we propose an Image disturbance (Image-D) strategy, which set the null-text noisy image as a clean image to extract additional details from it. Specifically, we empirically set $x_{\sigma} = x_{ref}$, i.e., the input image.
The results depicted in the second row on the right of Figure \ref{compare_ri} demonstrate that utilizing Image-D leads to enhanced preservation of intricate features present in the input image, and thereby results in superior fidelity.
The algorithm of image cartoonization with Back-D or Image-D is summarized in Algorithm~\ref{algo2}.

\begin{algorithm}[t]
	\caption{Free generation of the cartoon with Back-D} 
        \label{algo1}
	\renewcommand{\algorithmicrequire}{\textbf{Input:}}
	\renewcommand{\algorithmicensure}{\textbf{Output:}}
	\begin{algorithmic}[1]
		\REQUIRE A pre-trained Diffusion model $\epsilon_{\theta}(\cdot)$ with classifier-free guidance, prompt $p$, guidance scale $\gamma$, rollback step $b$ and disturbance time $s$.
		\ENSURE The cartoon-style image $x_0^*$.
            \STATE $x_T \sim \mathcal{N}(0,I)$
		\FOR {$t$ from $T$ to $0$ }
            \IF {t > s}
            \STATE Get $\epsilon_t$ via Eq.\eqref{cfg}
            \ELSE 
            \STATE  $\epsilon_{\theta}(x_{\sigma}|\oslash) = \epsilon_{\theta}(x_{t+b}|\oslash)$.
            \STATE $\epsilon_t =\epsilon_\theta (x_{\sigma} | \oslash )+ \gamma ( \epsilon_{\theta}(x_t| p) - \epsilon_\theta (x_{\sigma} | \oslash ) )$
            \ENDIF
            \STATE $x_{t-1} \leftarrow \epsilon_t$ via Eq.\eqref{x_t}.
		\ENDFOR
		\RETURN $x_0$ as $x_0^*$.
	\end{algorithmic}
\end{algorithm}

\begin{figure}[t]
  \centering
  \includegraphics[width=\linewidth]{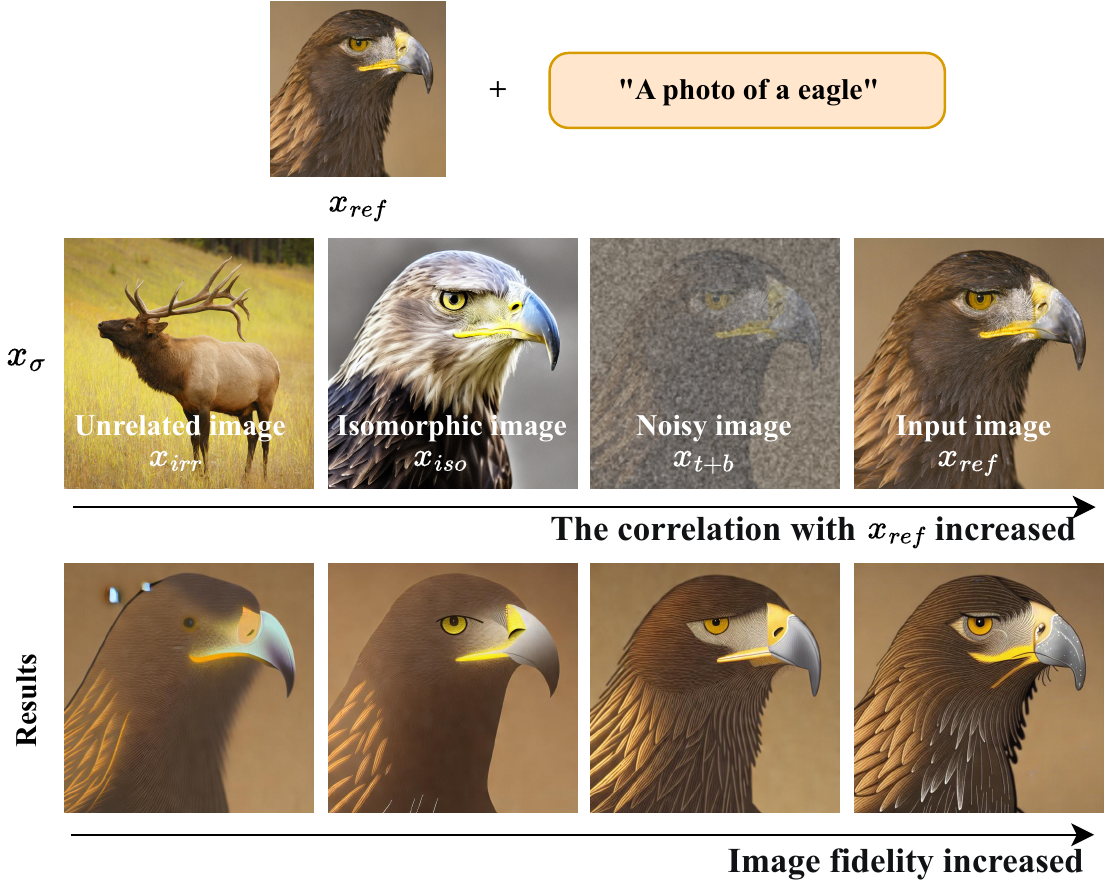}
  \vspace{-2.3em}
  \caption{The cartoon effect is affected by the correlation between the noise input of null-text and the input image. 
  }
  \label{fig_analysis}
\end{figure}

\begin{algorithm}[t]
	\caption{Image cartoonization based on noise diturance} 
        \label{algo2}
	\renewcommand{\algorithmicrequire}{\textbf{Input:}}
	\renewcommand{\algorithmicensure}{\textbf{Output:}}
	\begin{algorithmic}[1]
		\REQUIRE A pre-trained Diffusion model $\epsilon_{\theta}(\cdot)$ with classifier-free guidance, prompt $p$, guidance scale $\gamma$, rollback step $b$, disturbance time $s$ and a specific input image $x_{ref}$.
		\ENSURE The cartoon-style image $x_0^*$.
            \STATE Adding noise for $x_{ref}$ with $t=s$ steps and get the noised image $x_s$.
		  \FOR {$t$ from $s$ to $0$ }
            \STATE $\epsilon_{\theta}(x_{\sigma}|\oslash) = \epsilon_{\theta}(x_{t+b}|\oslash)$ if use Back-D else $\epsilon_{\theta}(x_{ref}|\oslash)$
            \STATE $\epsilon_t =\epsilon_\theta (x_{\sigma} | \oslash )+ \gamma ( \epsilon_{\theta}(x_t| p) - \epsilon_\theta (x_{\sigma} | \oslash ) )$
            \STATE $x_{t-1} \leftarrow \epsilon_t$ via Eq.\eqref{x_t}.
		\ENDFOR
		\RETURN $x_0$ as $x_0^*$.
	\end{algorithmic}
\end{algorithm}

\subsection{Analysis of the null-text noisy image}
\label{analysis}

The main idea of this work is to substitute the null-text noisy image with a disturbance image $x_{\sigma}$, which introduces misalignment between the null-text noisy image and text noisy image.
By utilizing $x_{t+b}$ and $x_{ref}$, Back-D and Image-D modify the sampling direction, thereby inducing cartoon-style generation. 

To explore the function of the null-text noisy image, we vary the correlation of $x_{\sigma}$ with $x_{ref}$.
Specifically, Figure~\ref{fig_analysis} illustrates two additional settings for $x_{\sigma}$, viz., an unrelated image $x_{irr}$ and an isotropic image $x_{iso}$ that shares structural similarity with $x_{ref}$. 
The degree of correlation between $x_{ref}$ and $x_{\sigma}$ in various settings satisfies: $x_{irr} < x_{iso} < x_{t+b} < x_{ref} (100\%).$
The results indicate that as the correlation degree between null-text noisy image and input images $x_{ref}$ increases, both the quality and fidelity of generation improve as showed in Figure~\ref{fig_analysis}. 

Notably, both Back-D and Image-D yield desirable cartoonization results, with the latter providing richer details. Furthermore, other experimental results (see supplementary material) demonstrate that Back-D and Image-D each have their respective advantages and can be used selectively depending on the specific case.


\begin{figure*}[t]
  \centering
  \includegraphics[width=\linewidth]{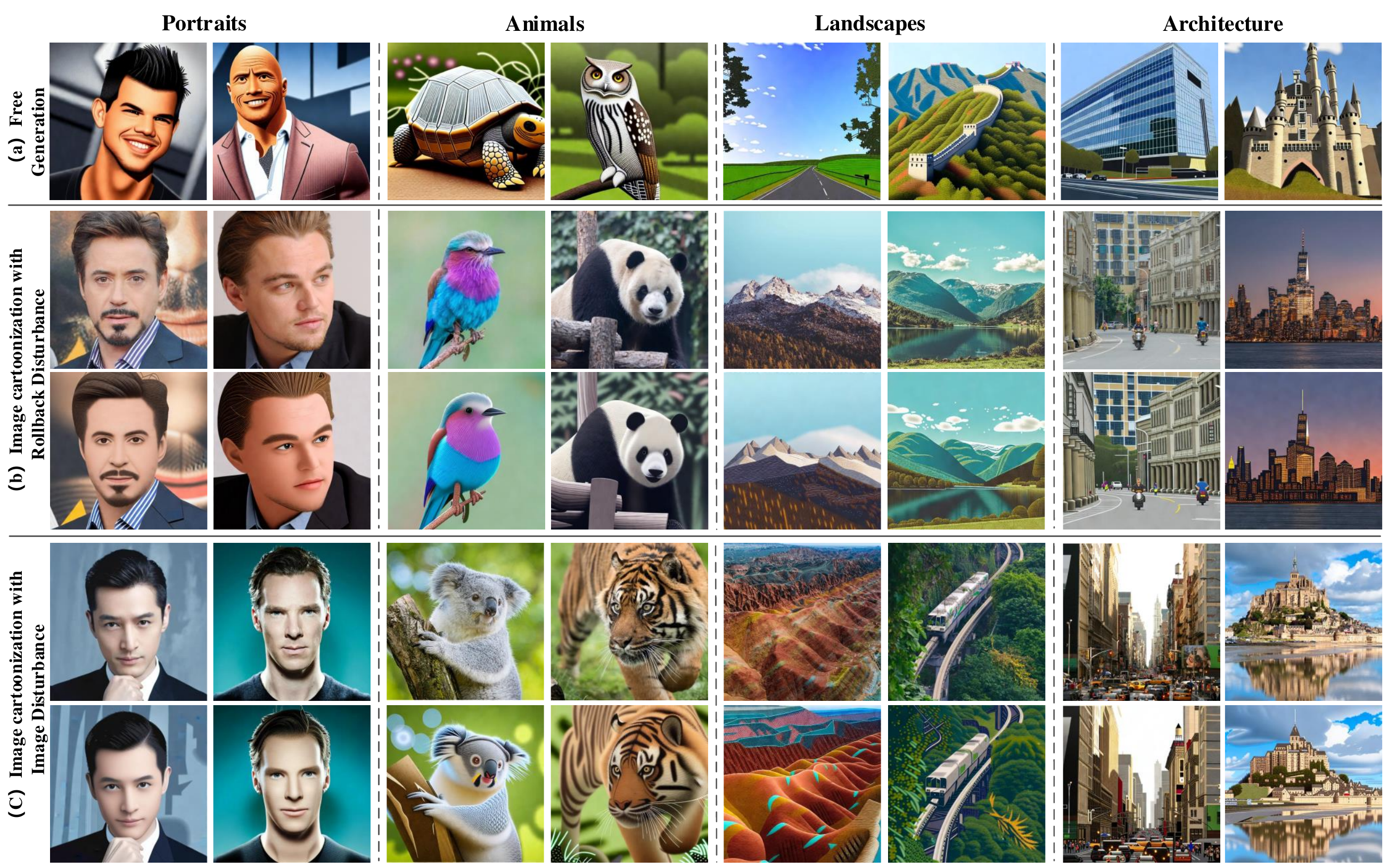}
  \vspace{-2.3em}
  \caption{Results of (a) free generation using Back-D, (b) Image cartoonization using Back-D, and (c) using Image-D. The results indicate that the proposed method enables free cartoon generation of portraits, animals, landscapes, and architectures while achieving image cartoonization.
  }
  \label{results_all}
\end{figure*}

\begin{figure*}[t]
  \centering
  \includegraphics[width=\linewidth]{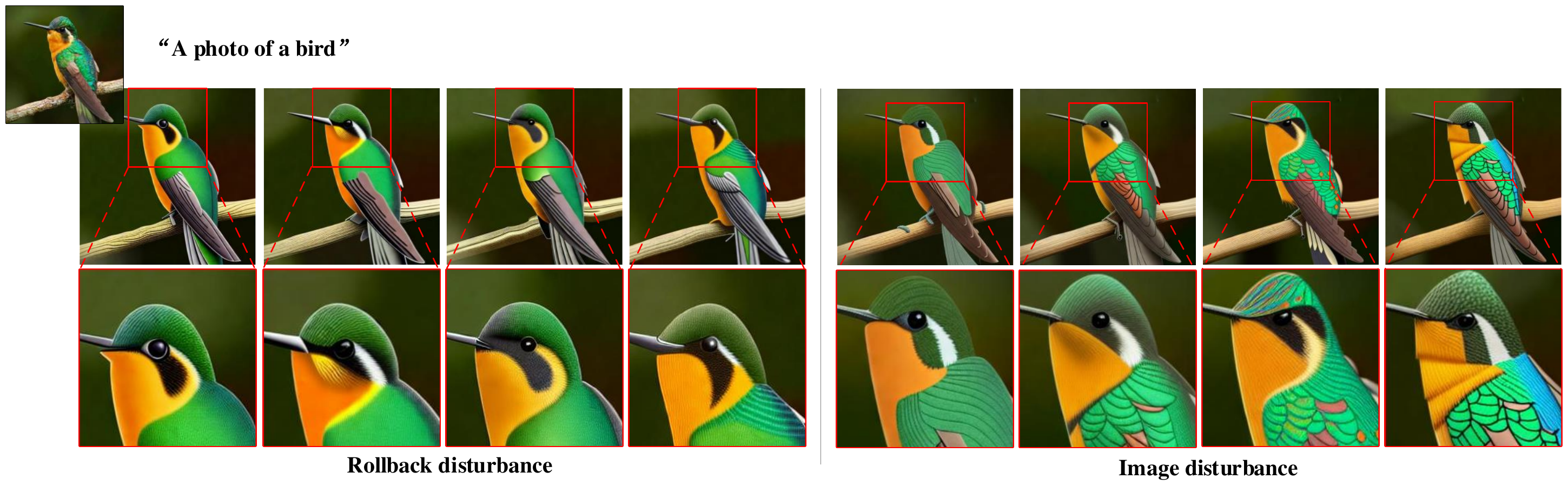}
 \vspace{-2.5em}
  \caption{
  Image cartoonization showcases diversity. The Image disturbance (Image-D) contains richer diversity of details.
  }
  \label{diversity}
\end{figure*}

\begin{figure*}[t]
  \centering
  \includegraphics[width=\linewidth]{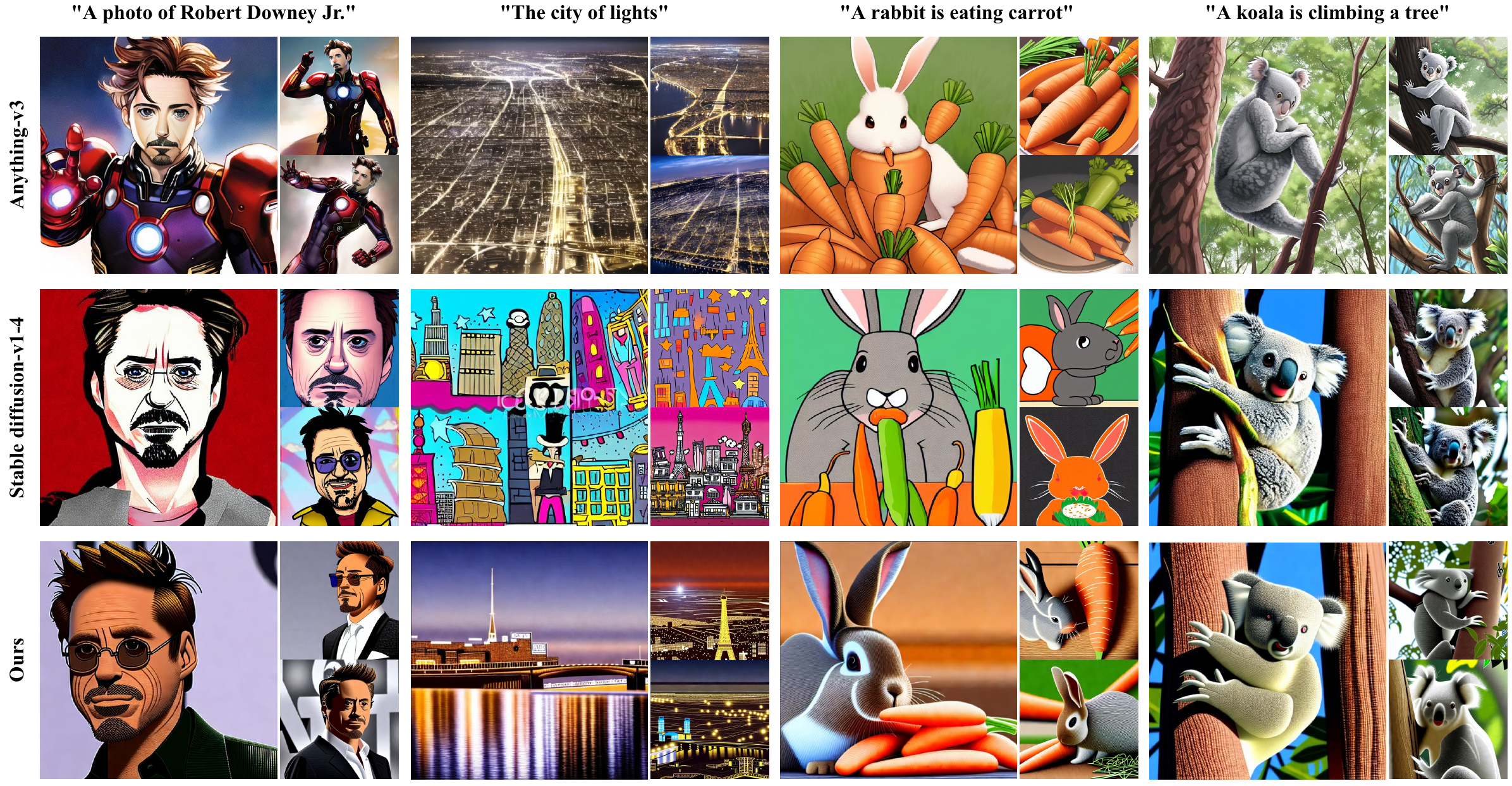}
  \vspace{-2.3em}	
  \caption{Comparison with other cartoon generation works. Our free generation using Back-D generates more accurate, vivid, and artistically textured cartoon images.}
  \label{compare1_main}
\end{figure*}

\begin{figure*}[t]
  \centering
  \includegraphics[width=\linewidth]{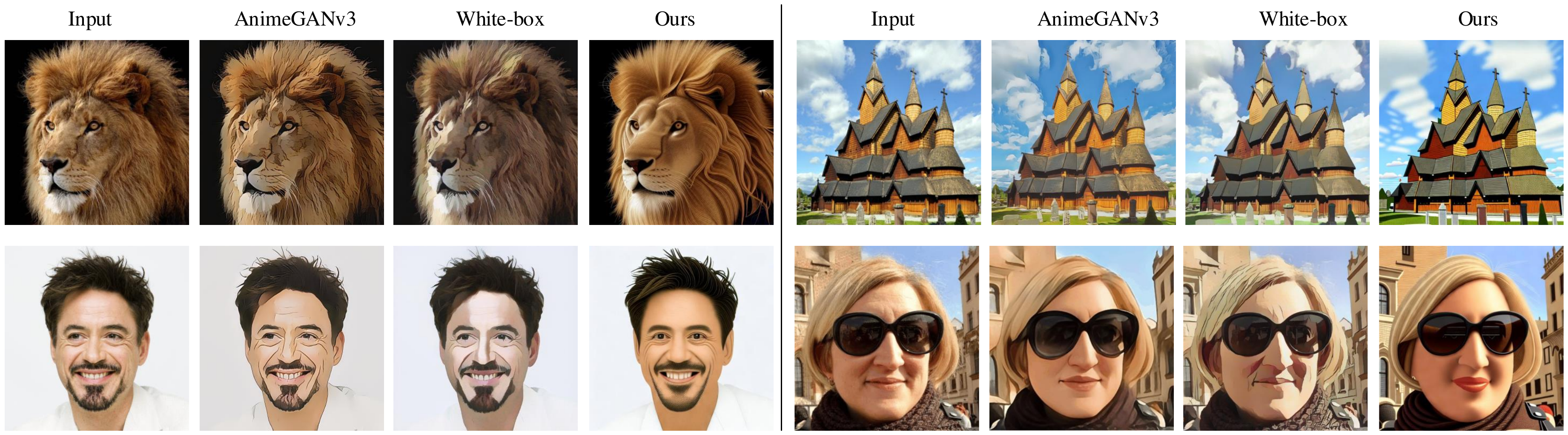}
  \vspace{-2.3em}	
  \caption{Comparison with other Image cartoonization works. 
  Our method produces cartoonized images that are more vivid and lifelike, approaching the three-dimensional quality of animated scenes.}
  \label{compare2_main}
\end{figure*}

\begin{figure*}[t]
  \centering
  \includegraphics[width=\linewidth]{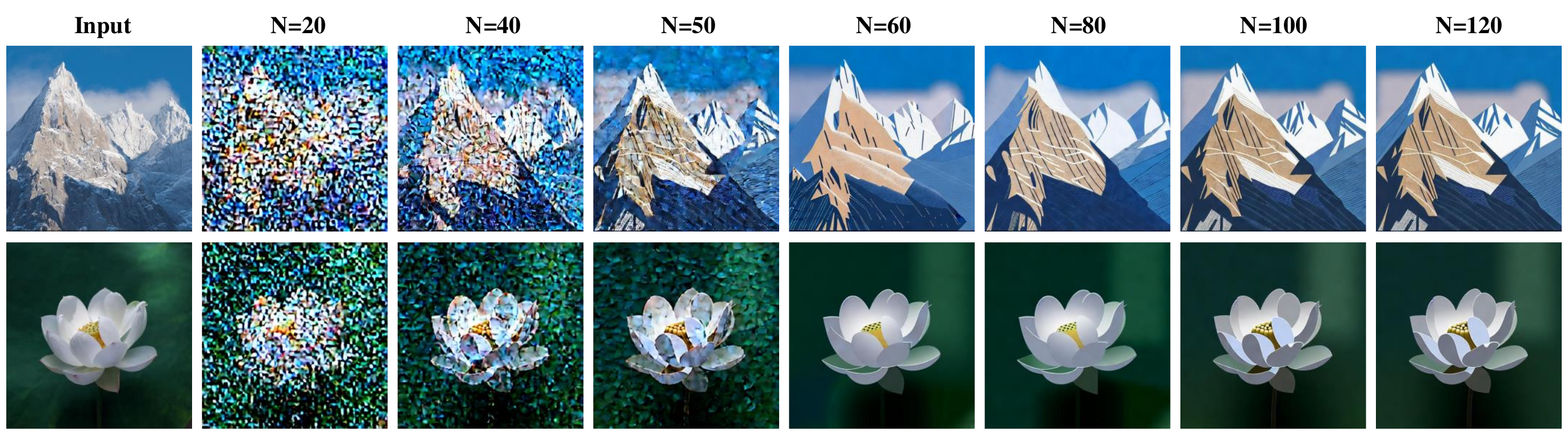}
  \vspace{-2.3em}	
  \caption{Study on the number of DDIM sampling steps $N$. $N$ larger than 60 yields a clean cartoon.
 }
    \label{study_ddim}
\end{figure*}

\begin{figure}[h]
  \centering
  \includegraphics[width=\linewidth]{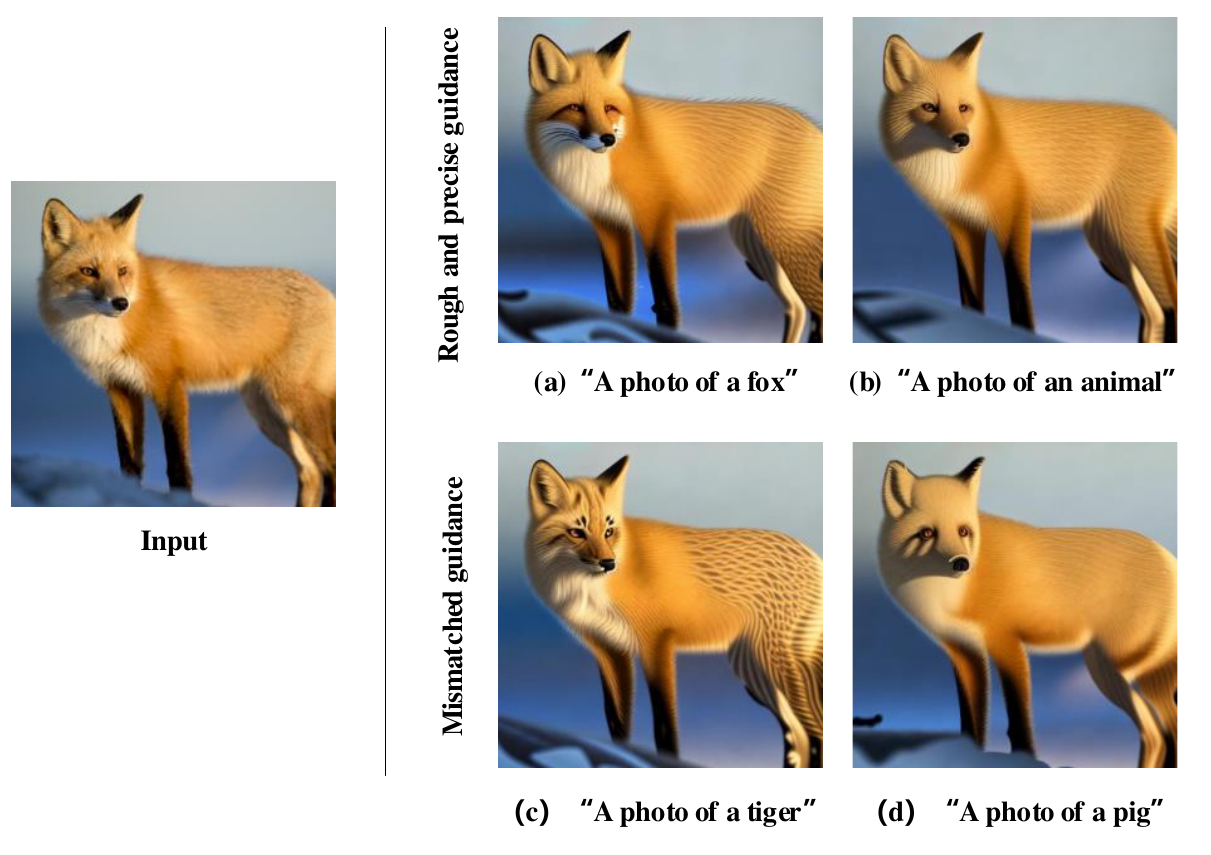}
  \vspace{-2.3em}	
  \caption{The influence of text guidance. 
  }
  \label{study_prompt}
\end{figure}

\begin{figure}[t]
  \centering
  \includegraphics[width=\linewidth]{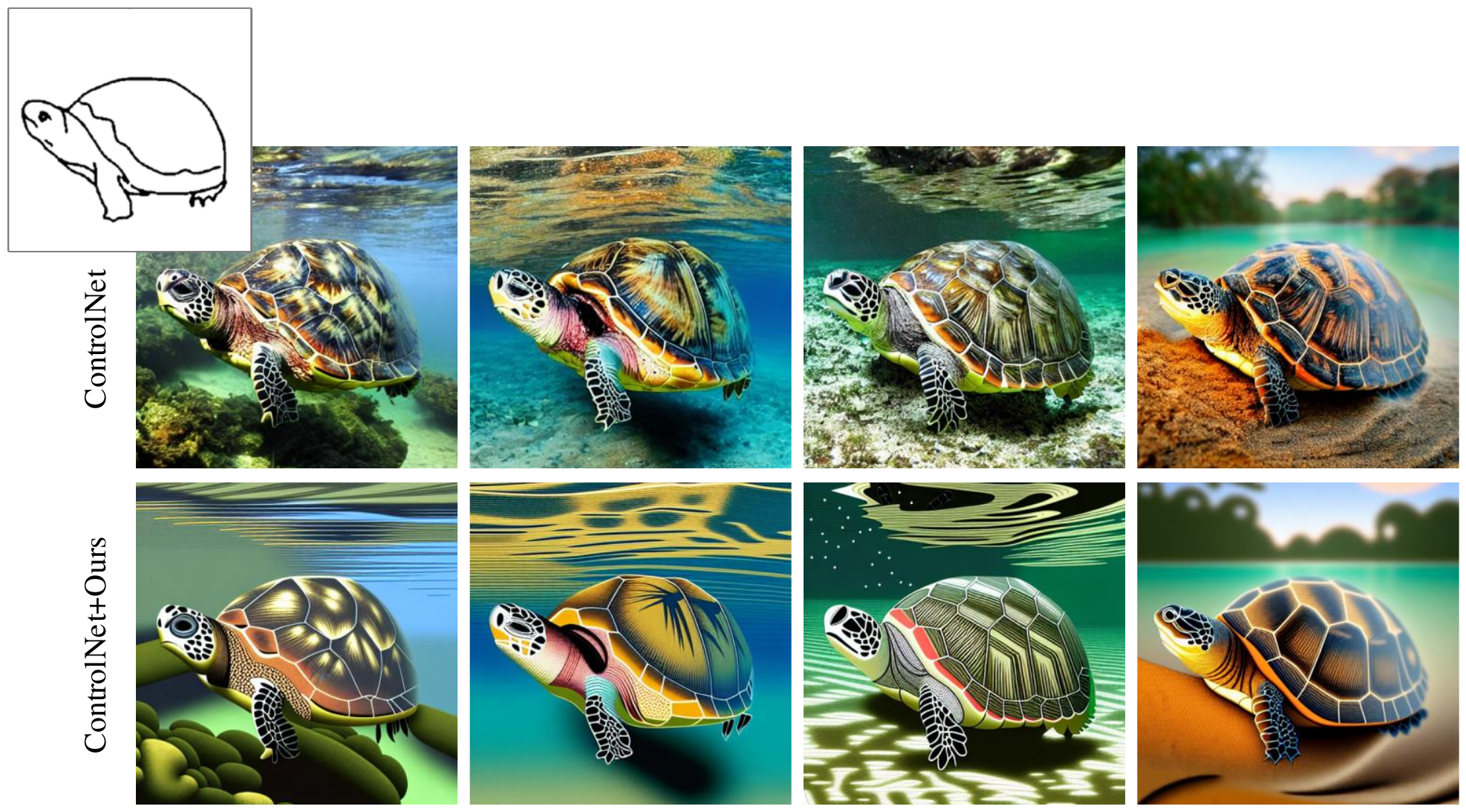}
  \vspace{-2.3em}	
  \caption{Application on ControlNet. 
  }
  \label{controlnet_main}
\end{figure}

\section{Experiments}
\label{experiments}
\subsection{Implementation details}

Our experiments are carried out based on Stableb Diffusion Model v1.4~\cite{RombachBLEO22} with DDIM steps initialized to 100. During most of our experimental trials, we assigned hyper-parameters $s\in (200, 300) $ and $b\in (200, 300)$. To achieve better cartoonization of special input, we suggest fine-tuning $s$ and $b$ without going over 400. The isomorphic image $x_{iso}$ in section 3.5 is provided by ControlNet~\cite{ControlNet23}.

\subsection{Results and Comparison}

The experimental results for three settings, i.e., free generation with Back-D, Image cartoonization with Back-D, and Image cartoonization with Image-D, are showcased in Figure~\ref{results_all}, which fully demonstrate the validity of our method. Check out the supplementary materials for more test results. It is worth mentioning that our method is training-free and serves as a plug-and-play component, its basic generation capability relies on the underlying model (Stable diffusion v1.4) it is deployed with. If the generative capability of the underlying model is improved, we can expect a significant enhancement in the cartoonization effect as well.

\textbf{The diversity of image cartoonization.} 
Exploiting the creative potential of text-guided diffusion models, our image cartoonization method based on noise disturbance yields a diverse array of results dependent upon input images.
Figure~\ref{diversity} illustrates the richness of the output.

\textbf{The impact of the guidance scale $\gamma$. } The guidance scale $\gamma$ plays a crucial role in classifier-free guidance. Through experimentation, we have discovered that, 
for $\gamma$ values within the range [2, 8], the degree of cartoonization increases with an increment in $\gamma$. Within the range [8, 12], stable and well-cartoonized graphics can be generated. However, if $\gamma$ exceeds 12, the generated image gradually incorporates more and more noise. Within the effective range of $\gamma$ for generating stable and well-transformed cartoon images, which is [8, 12], the optimal cartoonized image is achieved when both $b$ and $s$ fall within the range of [200, 300].

\textbf{Comparison in a free generation.} Figure~\ref{compare1_main} displays a comparison between our method and cartoon image generation model Anything v3~\cite{RombachBLEO22} and stable diffusion model v1.4\cite{RombachBLEO22}. Anything v3 is trained extensively with cartoon images but fails to accurately generate cartoons for new concepts or scenes not featured within its training data-as seen. For example, the case "A Photo of Robert Downey Jr." and the case "The City of Lights". 
It also suffers from scenario construction failure (case "A rabbit is eating carrot") and over-anthropomorphization of animals (case "A koala is climbing a tree""A koala is climbing a tree").
Meanwhile, the stable diffusion model v1.4 operates by modifying guided prompts "xxx" with "xxx in cartoon style", and its resulting images lack spatial information, as demonstrated in the first three cases of row 2 in Figure~\ref{compare1_main}.
Conversely, our method generates more accurate, vivid, and artistically textured cartoon images.

\textbf{Comparison in image cartoonization.}
We compare our cartoonization method with two well-known established techniques, i.e., AnimeGANv3~\cite{ChenLC19} and white-box~\cite{WangY20}, and the experimental results are presented in Figure~\ref{compare2_main}.
As can be seen from the comparison, 
both AnimeGANv3 and White-box apply additional line textures for planarization purposes, resulting in content being divided into blocks and appearing more akin to comic book illustrations.
By contrast, the results achieved through our Back-D are markedly distinct from previous efforts, and more vivid and lifelike, resulting in a style that is better suited for animation scenes.

\vspace{-0.8em}	
\subsection{The influence of text guidance}
In comparison to other image cartoonization methods that transform an input image solely based on its visual content, our text-guided cartoonization method, which generates an image not only based on the original image but also the text prompt, offers an innovative basis for the diversity and creativity of generated cartoons.

\textbf{Rough and precise guidance.}
We conducted experiments testing the influence of rough and precise text guidance on creating cartoon-style images. 
Based on the picture of a fox on the left of Figure~\ref{study_prompt}, we created two cartoon versions of it through rough guidance "a photo of an animal" and precise guidance "a photo of a fox" respectively. 
The results shown in Figure~\ref{study_prompt} (b) demonstrate that the model produces a simplified cartoonization of the original image under rough guidance. However, when given accurate categorization information, the model could successfully infuse fox-specific characteristics into the cartoon version, such as the slender eyes and long beard in Figure~\ref{study_prompt} (a).
It is clear that our method represents an intriguing and expressive means of generating cartoons. With appropriate text guidance, the resulting cartoon effect can be highly captivating.

\textbf{Mismatched guidance.}
To investigate the influence of text guidance more extensively, we tested our cartoonization approach under the context of mismatched text prompts.
Figure~\ref{study_prompt} (c) and (d) present the results produced by setting prompts to "a photo of a tiger" and "a photo of a pig" respectively, despite using an image of a fox as input. 
The results demonstrate that 
the combination of input images and text prompts can serve as inspiration for creating more creative cartoon characters.

\subsection{Restrictions on the sampling steps}

We conducted research on the number of DDIM sampling steps $N$, which determines the actual number of samplings in the process of $T \rightarrow 0$. The actual number of sampling $k$ satisfies $k = \frac{s * N}{T}$,
where $s$ denotes the time step at which the execution of noise perturbation commences.
Figure~\ref{study_ddim} depicts the cartoonization results of $N$ at varying values, and furthermore, as $N$ increases, the noise in the resulting images progressively decreases.
The results shows that noise disturbance fails to establish a stable sampling direction and consequentially results in significant noise throughout the generated image when $N<60$. 

\subsection{Application on ControlNet}
As a plug-and-play cartoonization component, the proposed methods can be readily applied to the classifier-free guided diffusion model. In this study, we investigated the efficacy of the proposed method in ControlNet~\cite{ControlNet23}, a neural network structure designed to regulate pre-trained large diffusion models for accommodating more input conditions. Specifically, we leveraged the Back-D proposed in this work to cartoonize the results of the scribble-to-image task in ControlNet and present the findings in Figure~\ref{controlnet_main}.
The outcomes indicate that the proposed technique is easily adaptable to other tasks and produces a favorable cartoon effect. 

\section{Conclusions}
In this work, we made a significant discovery that null-text guidance in the diffusion model is secretly a cartoon-style creator. Specifically, by iteratively perturbing the null-text guidance with our proposed Rollback disturbance (Back-D) and Image disturbance (Image-D) strategies, we were able to generate cartoons effortlessly and cartoonize specific input images.
we systematically investigated and analyzed the impact of various hyper-parameters on the proposed methods, elucidating the appropriate conditions for effective cartoonization. 
Notably, our approach outperforms existing techniques in terms of generating precise, vivid, and diverse cartoons. Moreover, it is notable that our methods are training-free and easily integrable as a plug-and-play component into any classifier-free guided diffusion model.

\begin{acks}
This work was partially supported by the National Natural Science Foundation of China: No. 91948303-1, No. 61803375, No. 12002380, No. 62106278, No. 62101575, No. 61906210; the Postgraduate Scientific Research Innovation Project of Hunan Province: QL20210018; the National Key R\&D Program of China (2021ZD0140301)
\end{acks}

\clearpage
\bibliographystyle{ACM-Reference-Format}
\bibliography{sample-base2}

\clearpage
\appendix
\section{Results of free generation}
\begin{figure*}[h]
  \centering
  \includegraphics[width=\linewidth]{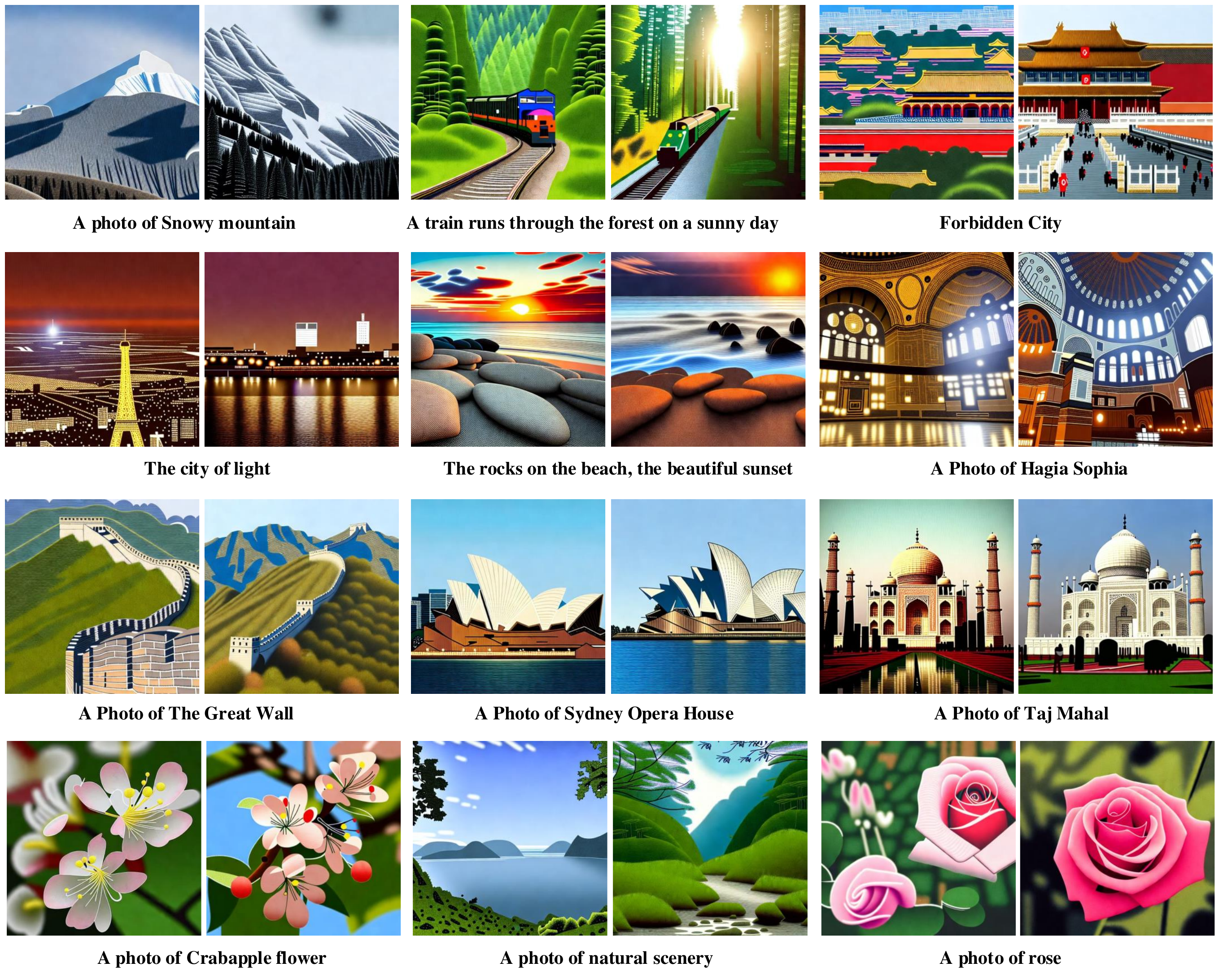}
  \caption{Free generation of cartoons using Rollback disturbance. The image is generated based on the textual prompts below. 
  }
  \label{mode1_1}
\end{figure*}
\begin{figure*}[h]
  \centering
  \includegraphics[width=\linewidth]{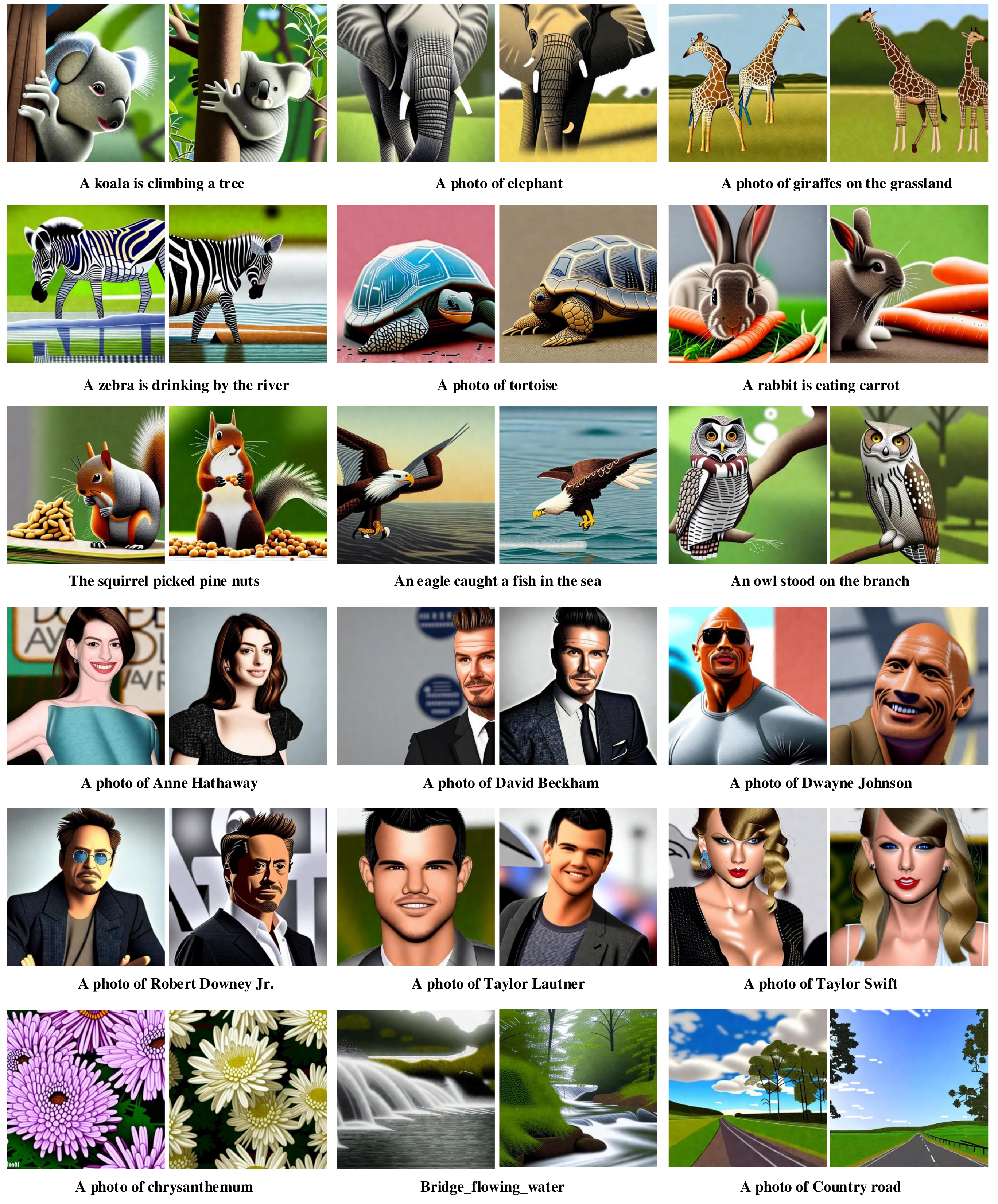}
  \caption{Free generation of cartoons using Rollback disturbance.  }
  \label{mode1_2}
\end{figure*}

We present additional examples of free generation in Figures~\ref{mode1_1} and \ref{mode1_2} as a supplement to Figure 6(a) in the main text. Free generation refers to the diffusion model generating cartoon images that are contextually relevant, by leveraging our proposed approach along with a given prompt.  Experimental results demonstrate that our method is effective in generating various types of cartoon images. Notably, the performance of free generation depends on the base model's ability to complete the text-to-image task. Our method simply applies a cartoonization process to the generated image based on the base model's output.

\section{Results of Image Cartoonization}
\begin{figure*}[h]
  \centering
  \includegraphics[width=\linewidth]{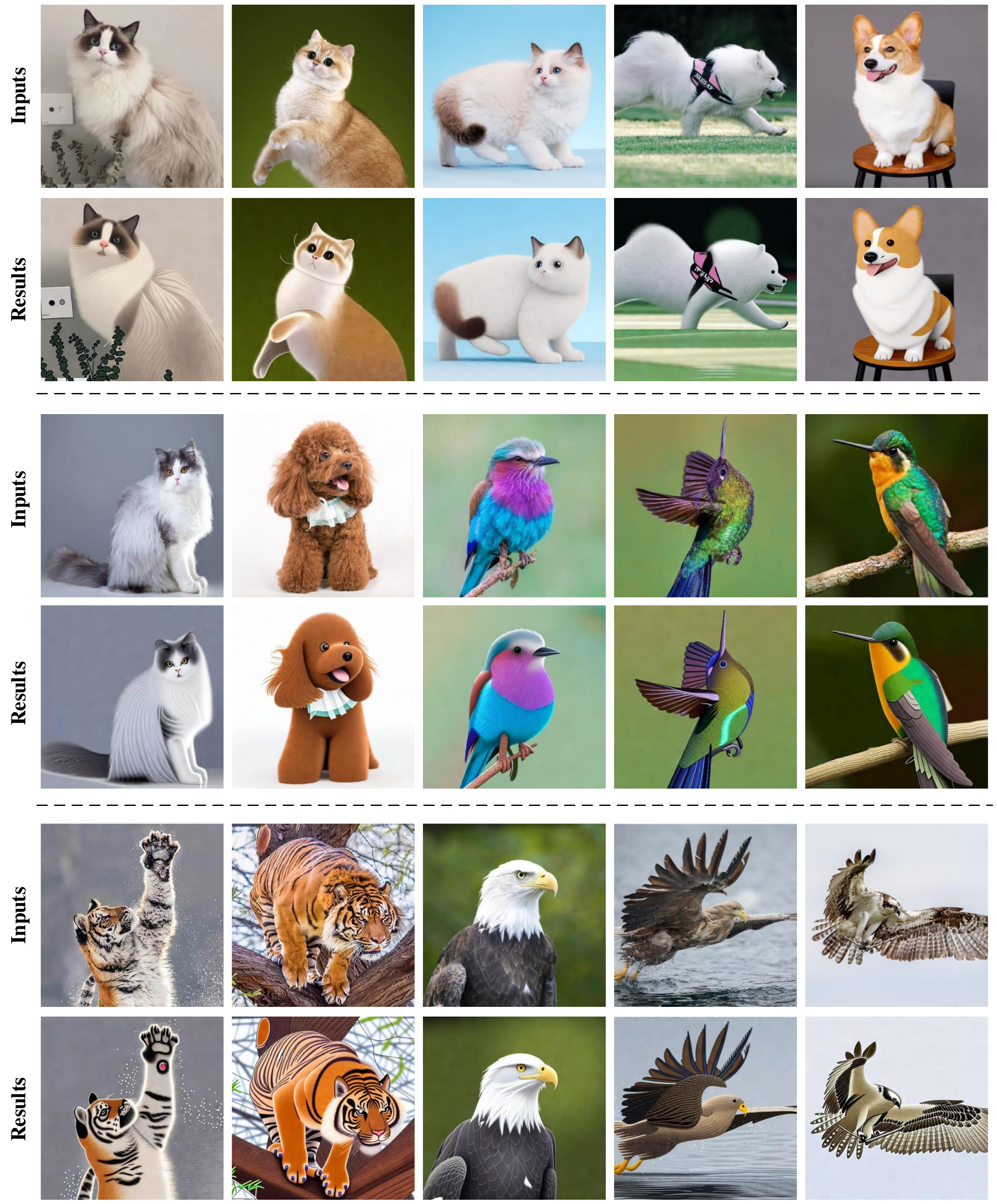}
  \caption{The results of Image cartoonization using Rollback disturbance.}
  \label{mode2_1}
\end{figure*}

\begin{figure*}[h]
  \centering
  \includegraphics[width=\linewidth]{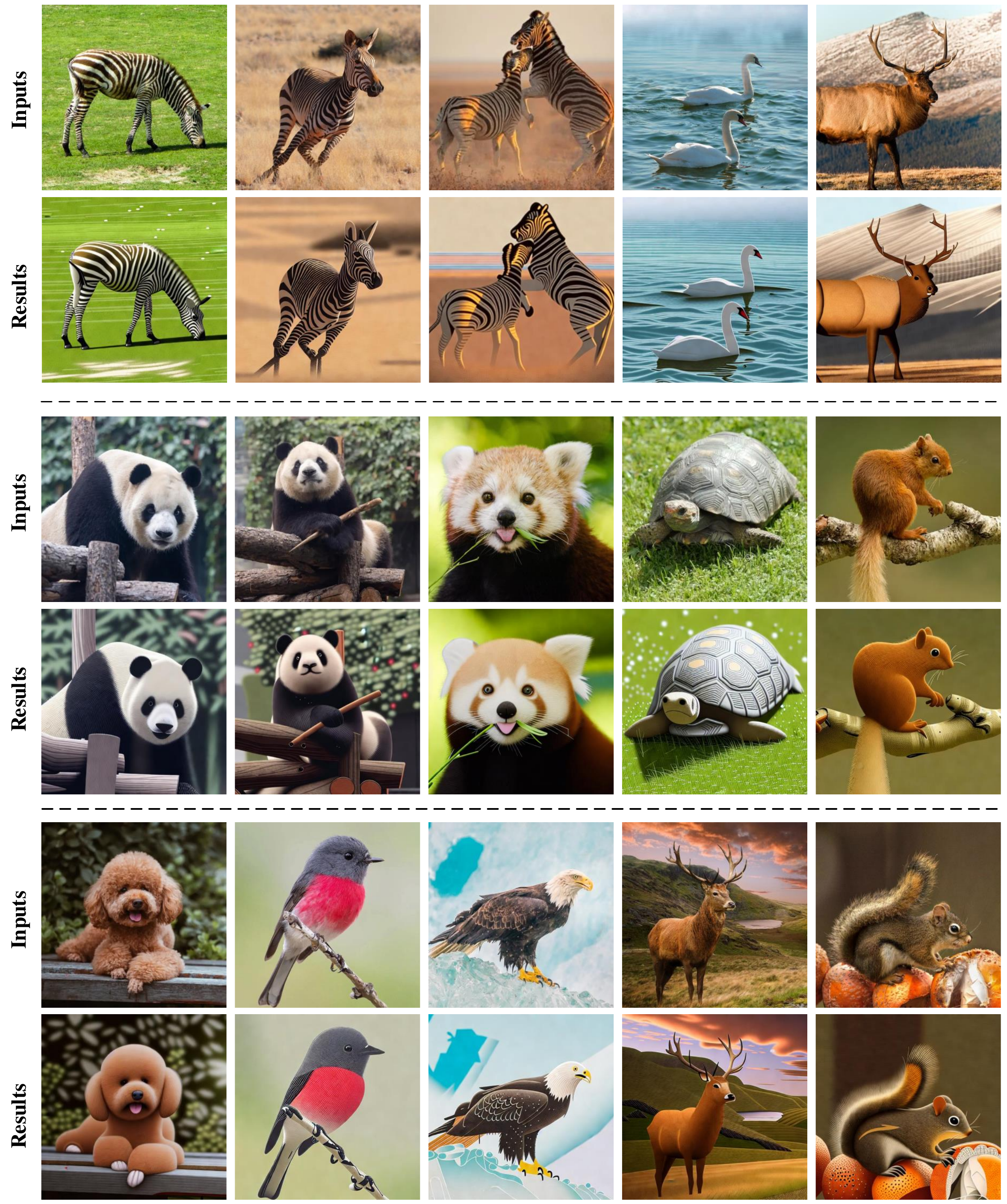}
  \caption{The results of Image cartoonization using Rollback disturbance. }
  \label{mode2_2}
\end{figure*}

\begin{figure*}[h]
  \centering
  \includegraphics[width=\linewidth]{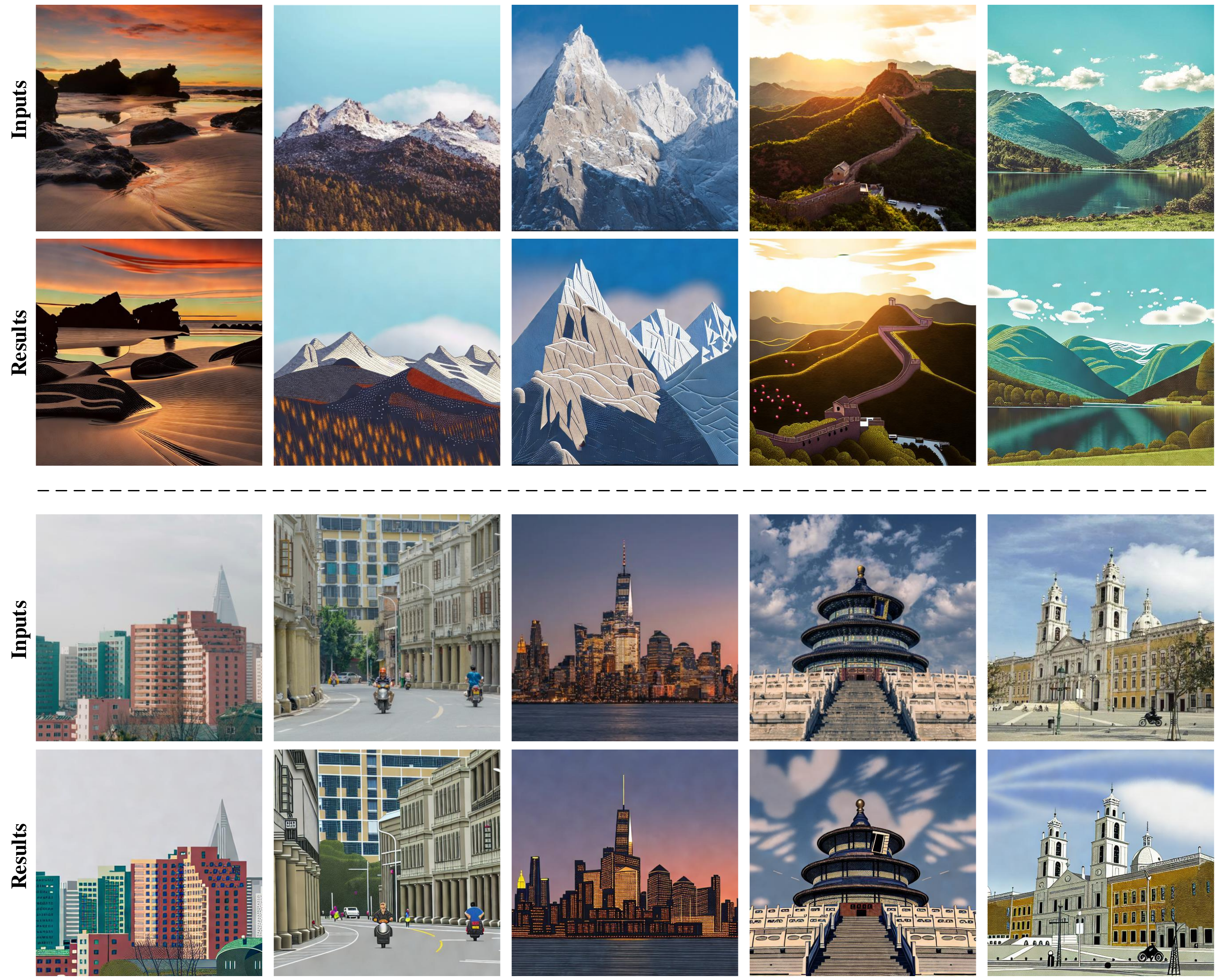}
  \caption{The results of Image cartoonization using Rollback disturbance.}
  \label{mode2_3}
\end{figure*}

\begin{figure*}[h]
  \centering
  \includegraphics[width=\linewidth]{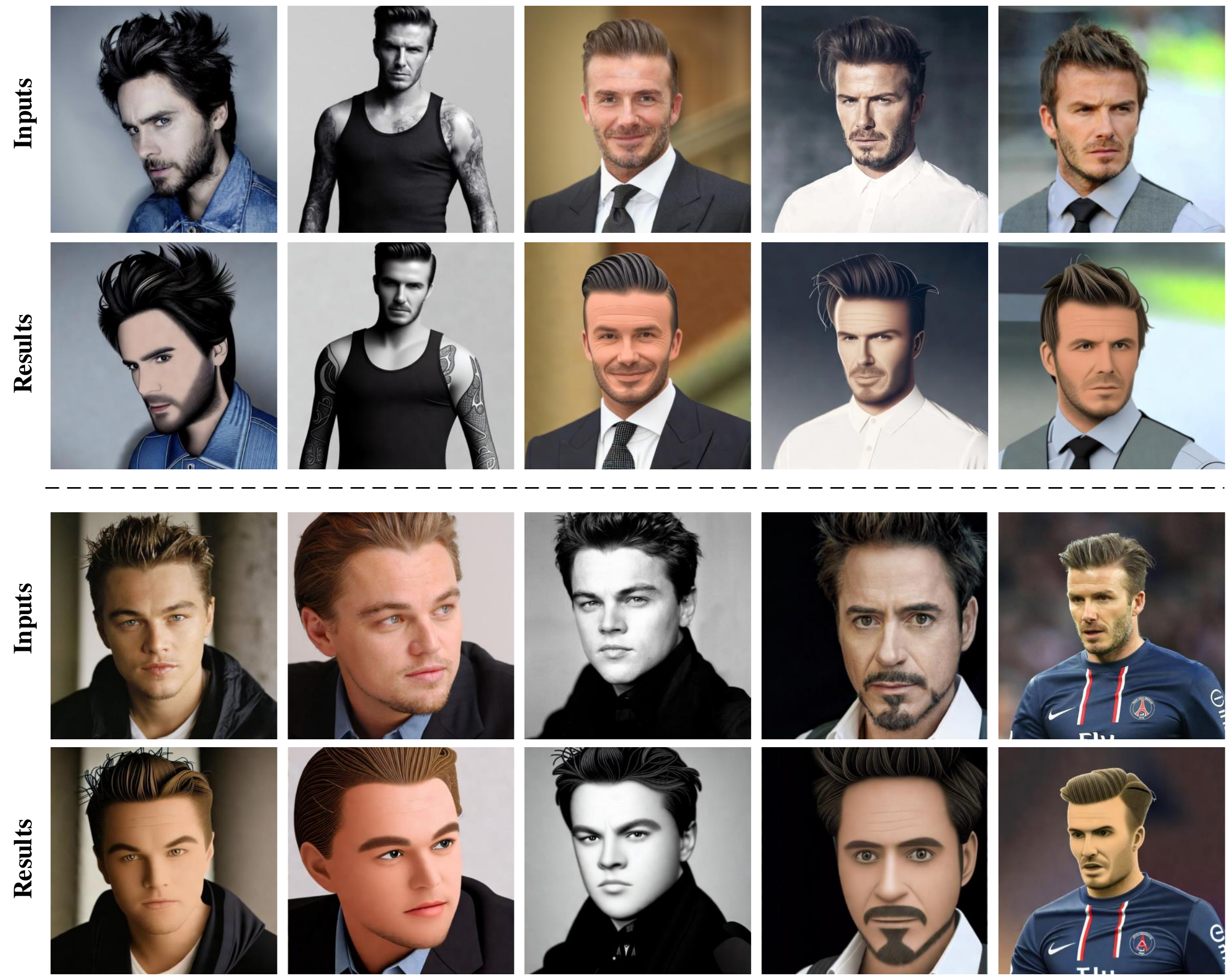}
  \caption{The results of Image cartoonization using Rollback disturbance.}
  \label{mode2_4}
\end{figure*}

\begin{figure*}[h]
  \centering
  \includegraphics[width=\linewidth]{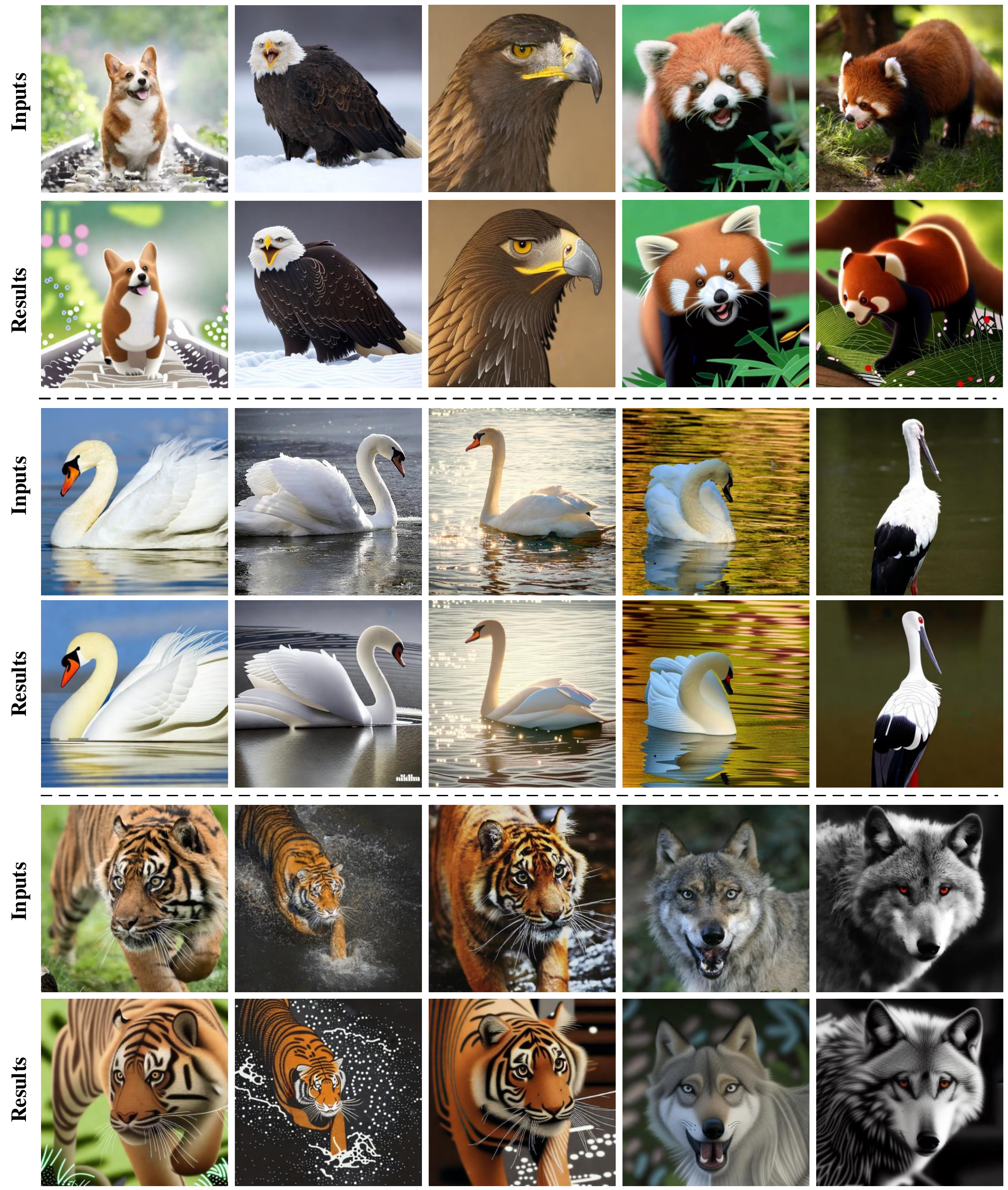}
  \caption{The results of Image cartoonization using Image disturbance. }
  \label{mode3_1}
\end{figure*}

\begin{figure*}[h]
  \centering
  \includegraphics[width=\linewidth]{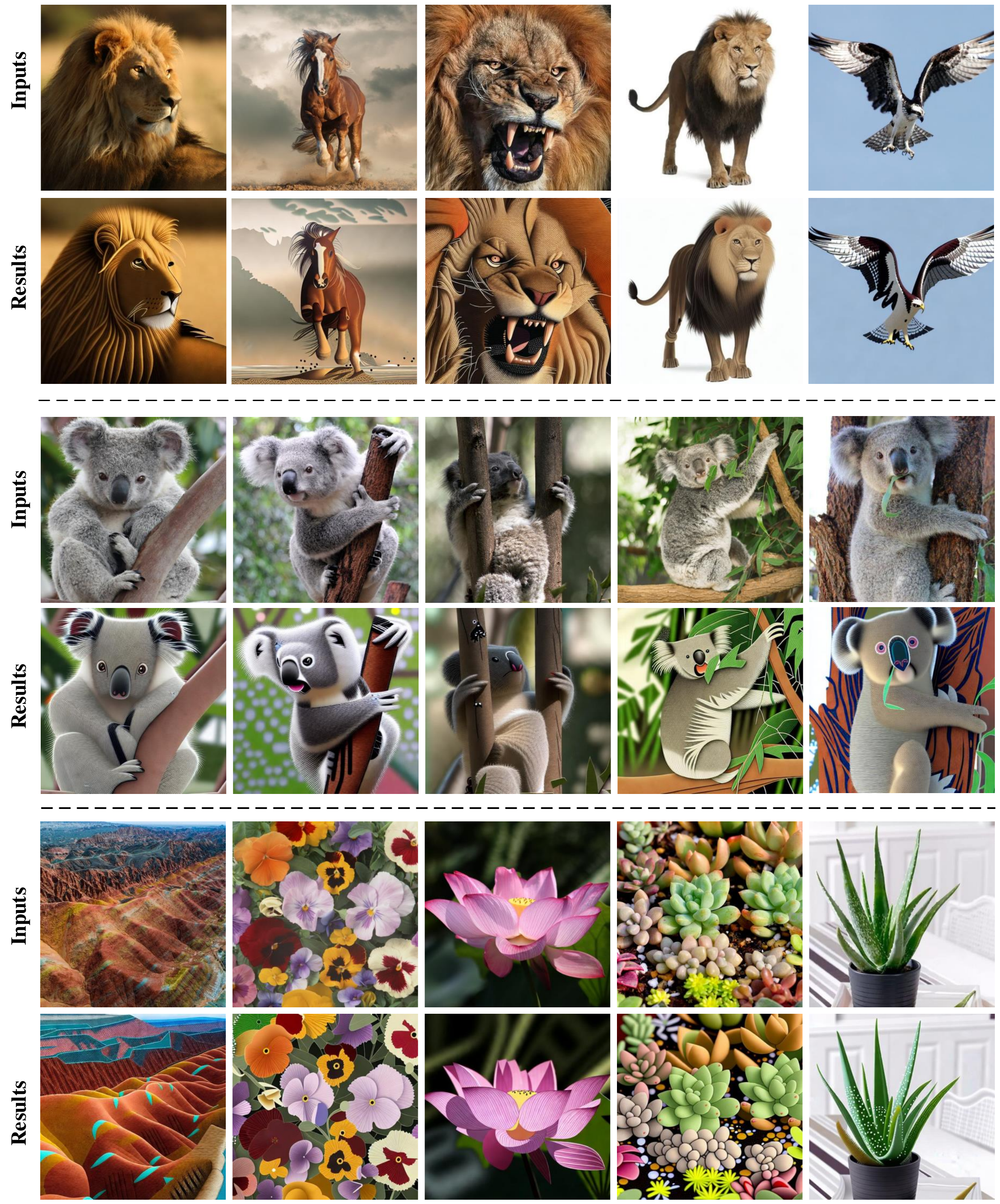}
  \caption{The results of Image cartoonization using Image disturbance.}
  \label{mode3_2}
\end{figure*}

\begin{figure*}[h]
  \centering
  \includegraphics[width=\linewidth]{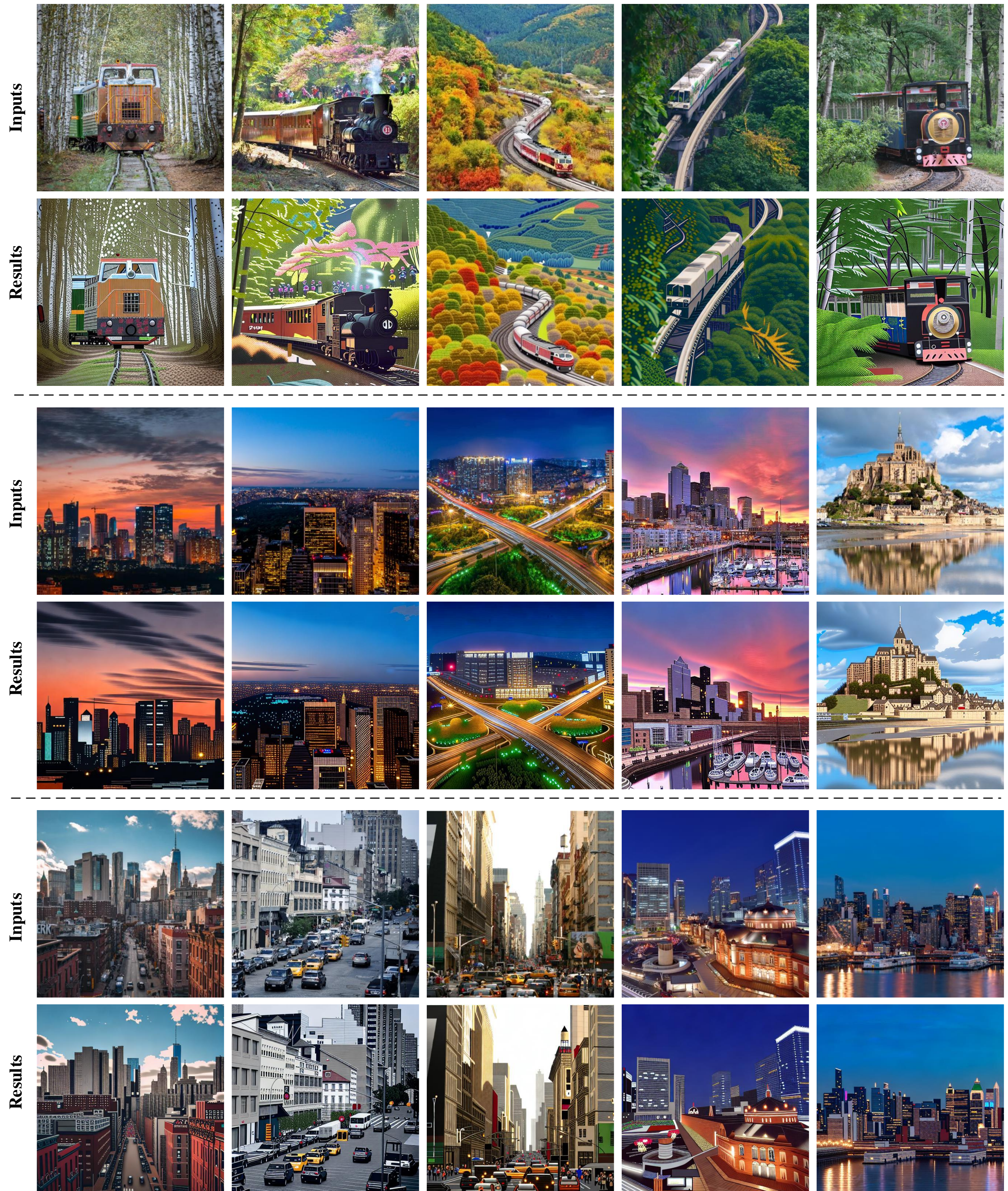}
  \caption{The results of Image cartoonization using Image disturbance.}
  \label{mode3_3}
\end{figure*}

To supplement Figure 6(b) in the main text, we present additional cases of image cartoonization using the Rollback disturbance technique in Figures ~\ref{mode2_1}, \ref{mode2_2}, \ref{mode2_3} and \ref{mode2_4}. Furthermore, to complement Figure 6(c) in the main text, Figures~\ref{mode3_1}, \ref{mode3_2}, and \ref{mode3_3} were used to introducing additional cases of image cartoonization generated via the Image disturbance technique.
Based on extensive testing, we have found that both Rollback disturbance and Image disturbance exhibit pronounced cartoonization effects. Overall, the former achieves a higher degree of cartoonization than the latter (at the expense of more detail being lost). However, we do not assert that the former is superior to the latter as a cartoonization technique. Our view is that Rollback disturbance and Image disturbance are suited to different types of input images, and users are free to choose between them based on their preferences for cartoonization outcomes. 

\begin{figure*}[h]
  \centering
  \includegraphics[width=\linewidth]{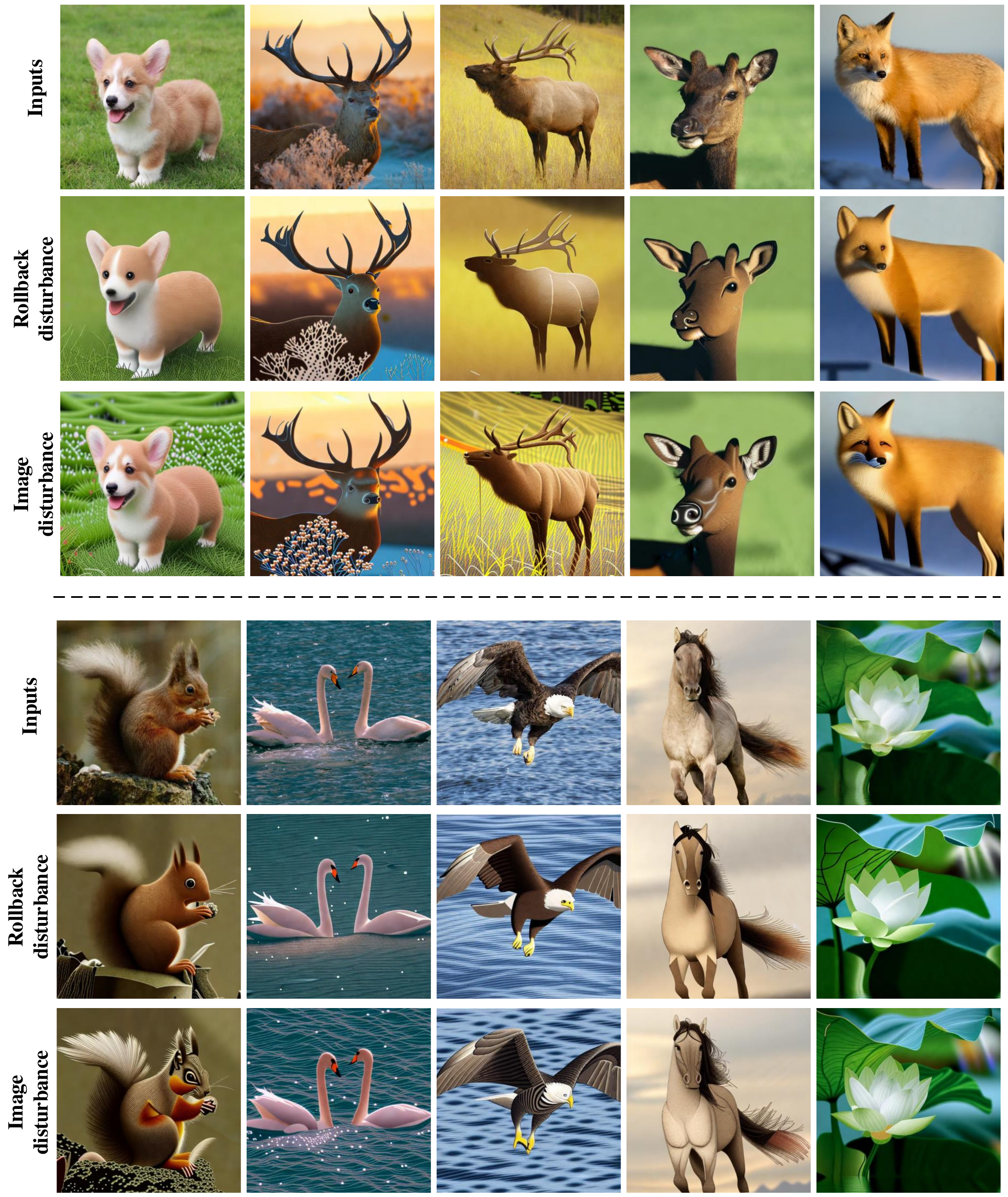}
  \caption{Comparison with Rollback disturbance and Image disturbance.}
  \label{compare1}
\end{figure*}

\begin{figure*}[h]
  \centering
  \includegraphics[width=\linewidth]{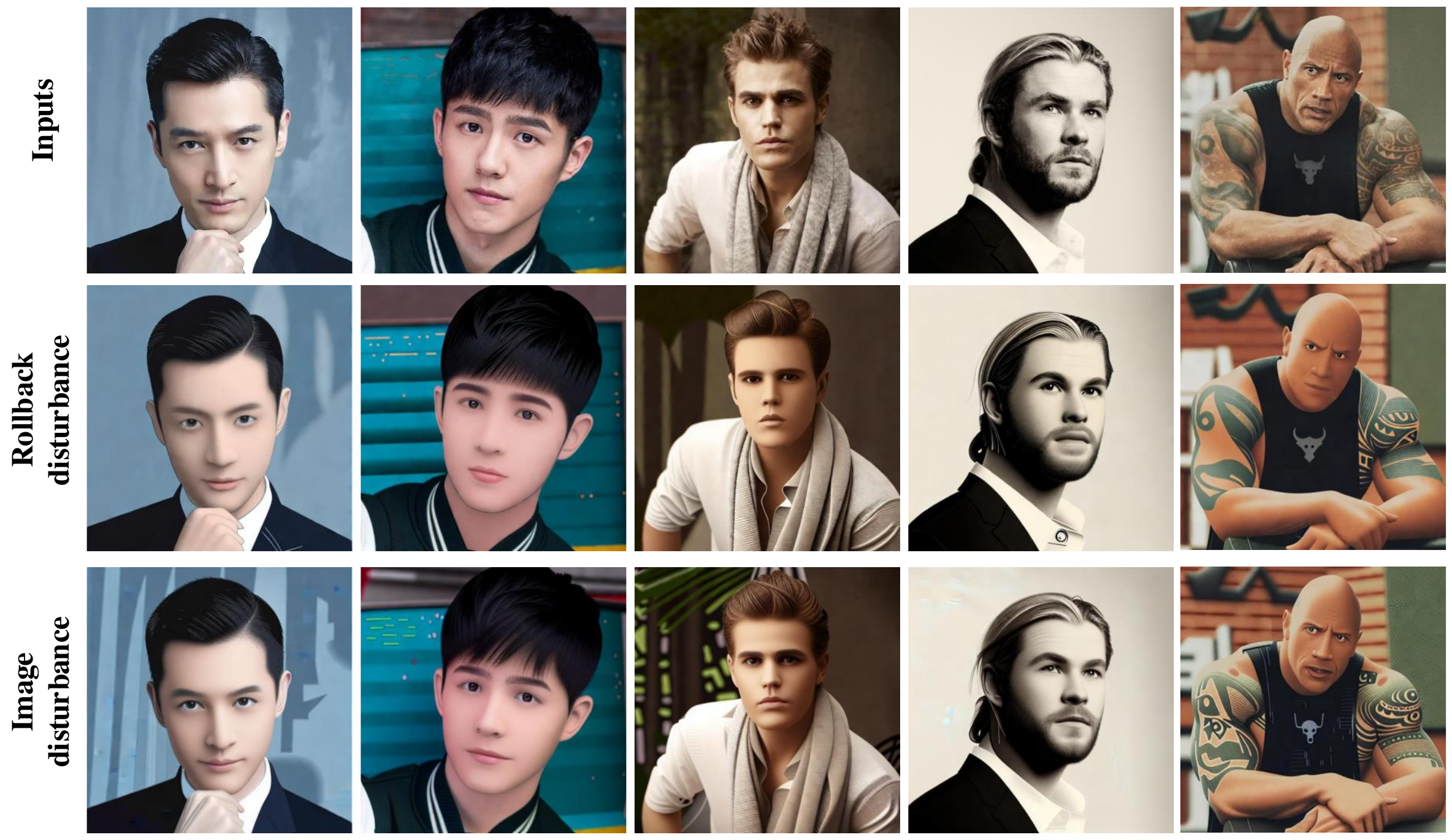}
  \caption{Comparison with Rollback disturbance and Image disturbance.}
  \label{compare2}
\end{figure*}

As demonstrated in Figures~\ref{compare1} and \ref{compare2}, which exhibit the cartoonized results of Rollback and Image disturbance, when generating cartoon-style images of animals and sceneries, Image disturbance enhances the expressiveness of the synthesized images by adding semantically meaningful details on top of the input image. Similarly, when it comes to producing cartoonized portraits, Image disturbance achieves high fidelity by capturing more detailed features of the input image.

\section{Diversity Exhibition}
\begin{figure*}[h]
  \centering
  \includegraphics[width=\linewidth]{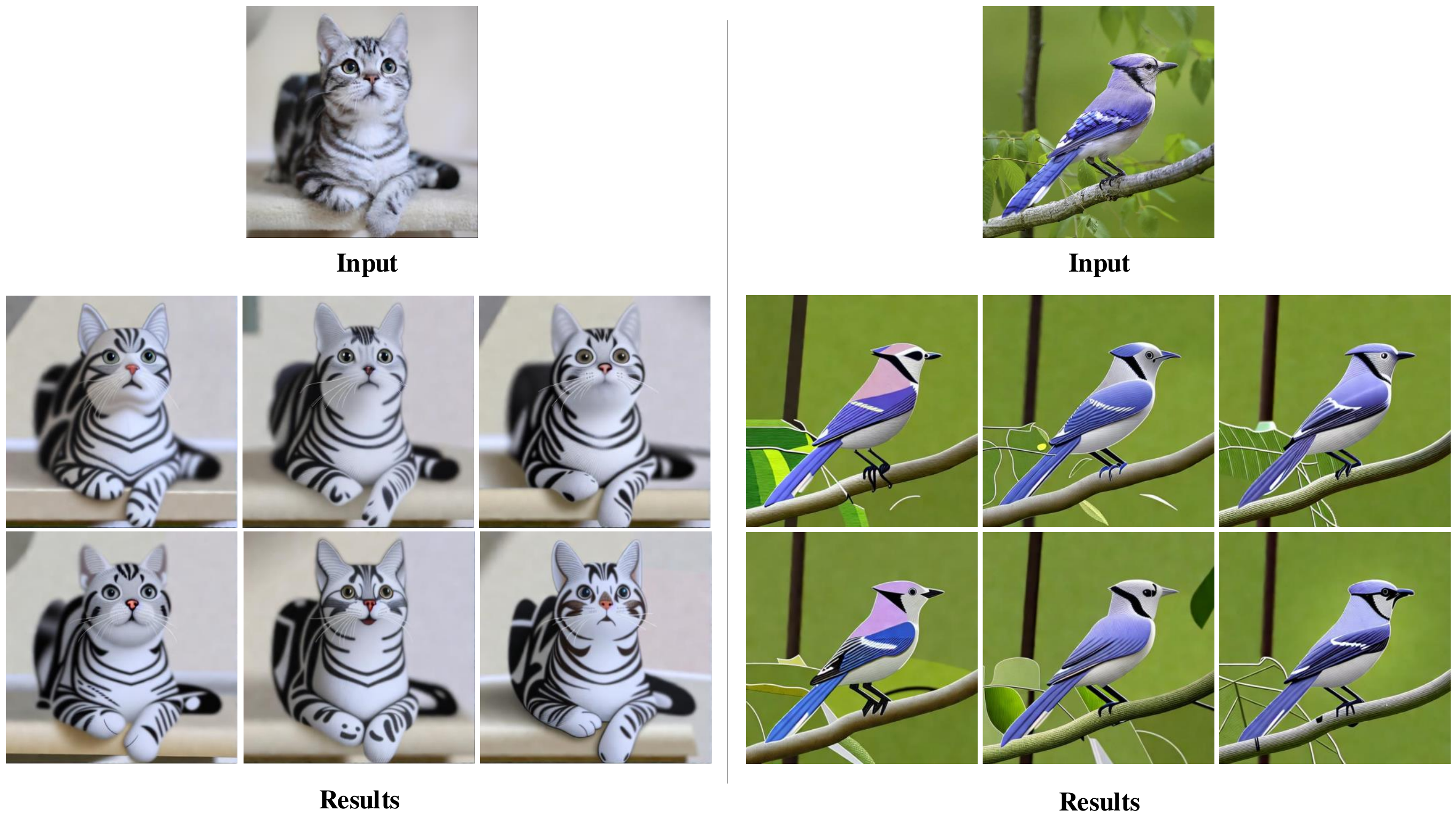}
  \caption{Diversity results demonstration based on Rollback disturbance.}
  \label{diversity_1}
\end{figure*}

\begin{figure*}[h]
  \centering
  \includegraphics[width=\linewidth]{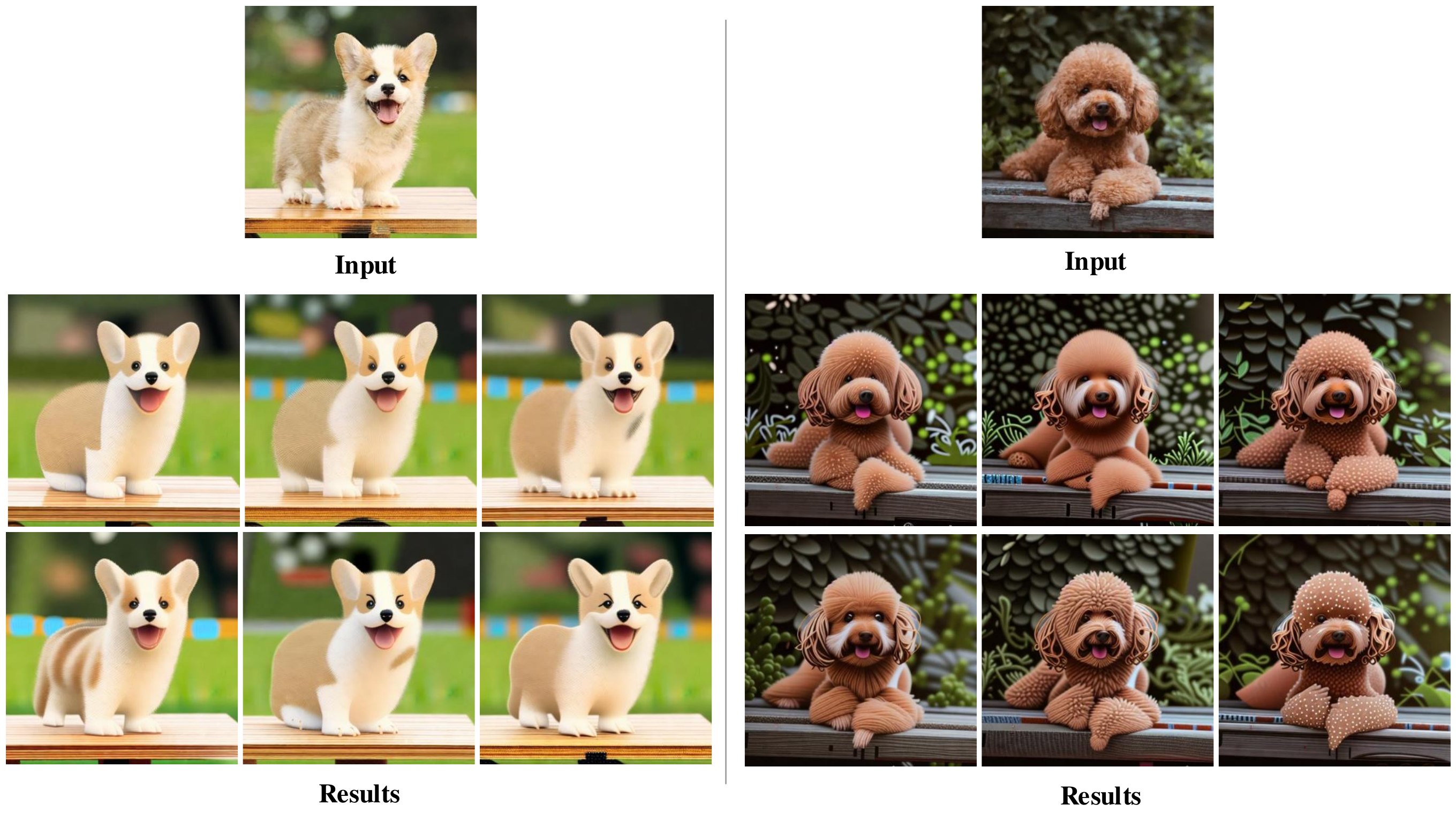}
  \caption{Diversity results demonstration based on Image disturbance.}
  \label{diversity_2}
\end{figure*}

We demonstrate the diversity of cartoonization outputs achieved through the usage of Rollback disturbance and Image disturbance in Figures~\ref{diversity_1} and \ref{diversity_2}, respectively. In contrast to conventional cartoonization techniques that enable a one-to-one mapping between input and cartoonized images, our approach enables creative cartoonization with a one-to-many mapping.

\section{Applications on ControlNet}
\begin{figure*}[h]
  \centering
  \includegraphics[width=\linewidth]{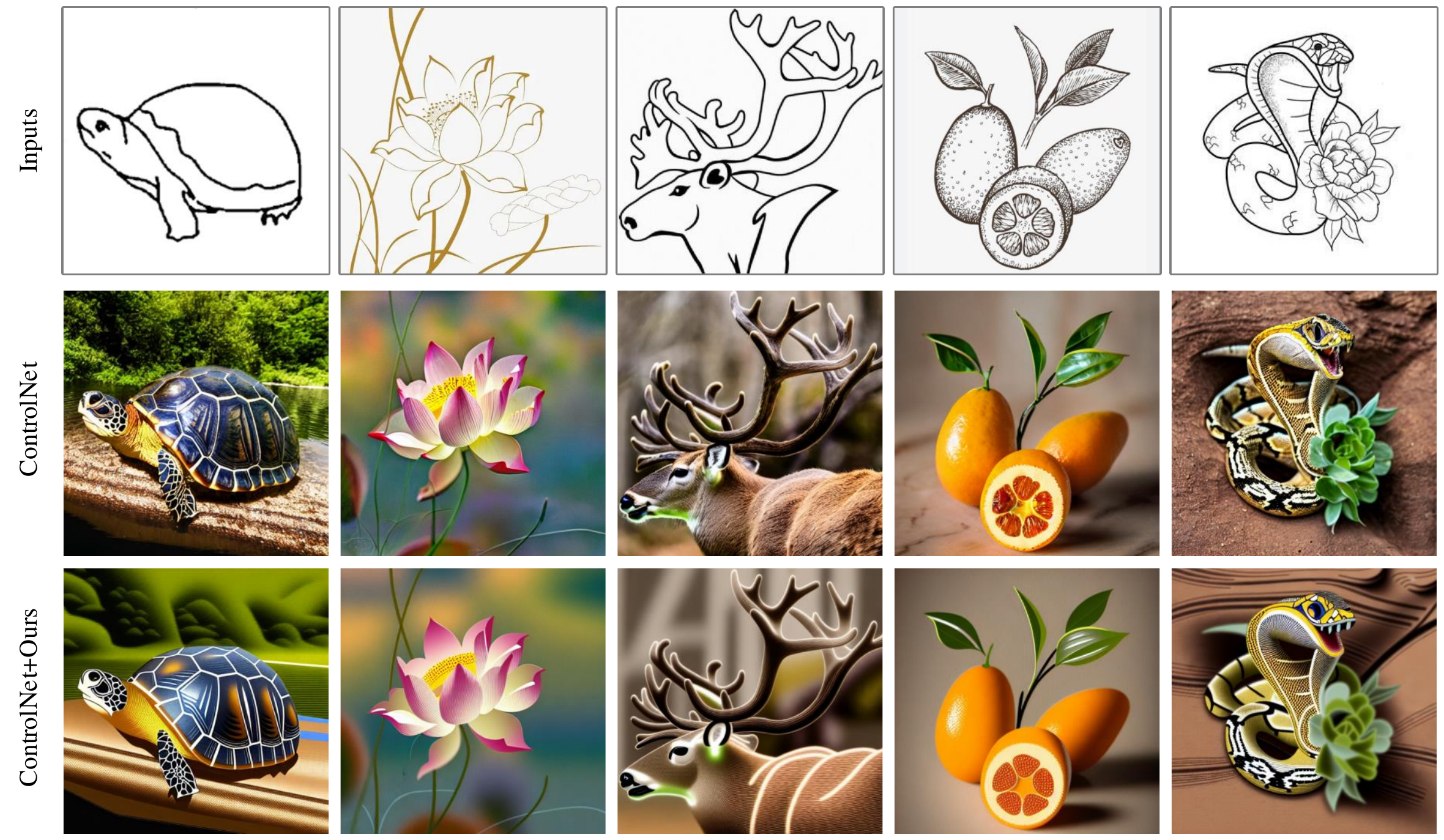}
  \caption{Applications on ControlNet}
  \label{controlnet}
\end{figure*}

Based on ControlNet\cite{ControlNet23}, we conducted further experiments on the scribble-to-image task and presented the results in Figure~\ref{controlnet}. The findings robustly demonstrate the efficacy of our proposed method as a plug-and-play cartoonization component that can be readily applied to various generative tasks.

\end{document}